%% file: ms.tex
\pdfoutput=1
\documentclass[11pt]{article}

\usepackage{acl}

\usepackage{times}
\usepackage{latexsym}
\usepackage{enumitem}
\usepackage{comment}

\usepackage[T1]{fontenc}
\usepackage[utf8]{inputenc}

\usepackage{microtype}

\usepackage[table,xcdraw]{xcolor}
\usepackage{lipsum}
\usepackage{amsmath}
\usepackage{amssymb}
\usepackage{scalerel,graphicx,xparse}
\usepackage{enumitem}
\usepackage{booktabs}
\usepackage{cancel}
\usepackage{csquotes}

\definecolor{violet-5}{RGB}{132, 94, 247}
\definecolor{darkgreen}{RGB}{0, 100, 0}

\NewDocumentCommand\crystalball{}{\includegraphics[height=.9em]{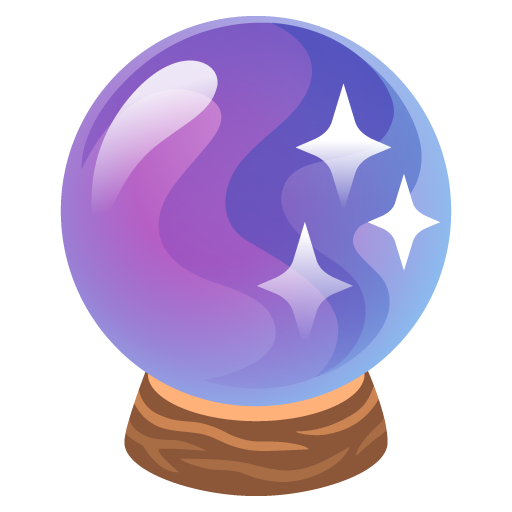}}
\NewDocumentCommand\explore{}{\includegraphics[height=.9em]{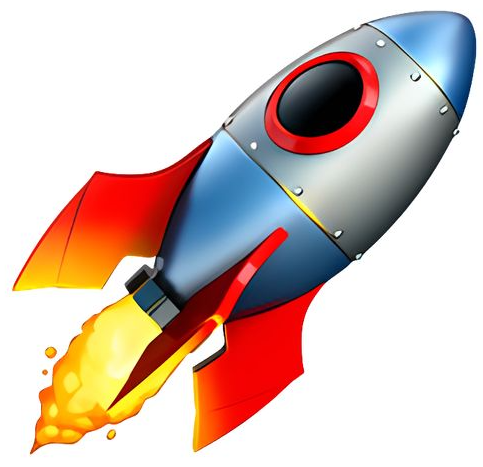}}
\NewDocumentCommand\exploit{}{\includegraphics[height=.9em]{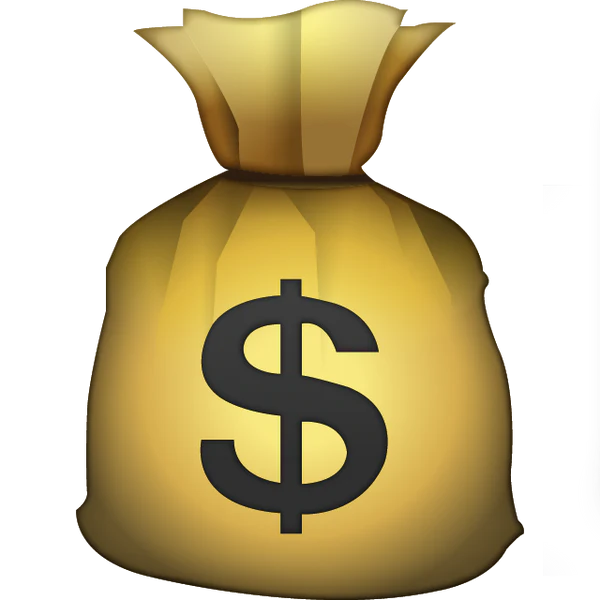}}
\NewDocumentCommand\interp{}{\includegraphics[height=.9em]{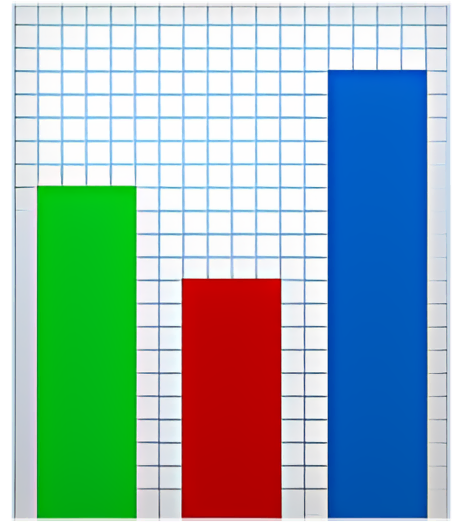}}
\NewDocumentCommand\llama{}{\includegraphics[height=.9em]{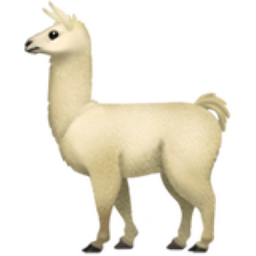}}
\NewDocumentCommand\zephyr{}{\includegraphics[height=.9em]{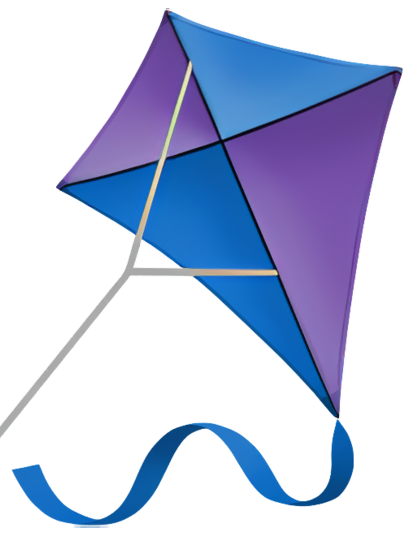}}

\NewDocumentCommand\scale{}{\includegraphics[height=.9em]{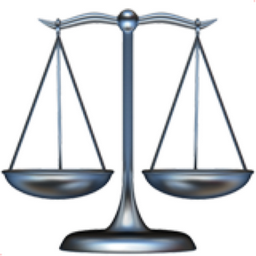}}
\NewDocumentCommand\methodname{}{\crystalball \texttt{\textbf{FortUne}} \texttt{\textbf{Dial}}}

\title{\textit{Deal, or no deal (or who knows)?}\\Forecasting Uncertainty in Conversations using Large Language Models}
\author{Anthony Sicilia$^\flat$\thanks{\ \ Work done on internship at Allen Institute for AI.} \quad Hyunwoo Kim$^{\natural}$ \quad Khyathi Raghavi Chandu$^\natural$  \\ \textbf{Malihe Alikhani}$^\flat$ \quad \textbf{Jack Hessel}$^\sharp$ \\
$^\flat$Northeastern University \quad $^\natural$Allen Institute for AI \quad $^\sharp$Samaya AI \\
\texttt{\{sicilia.a, m.alikhani\}@northeastern.edu} \\ \texttt{\{hyunwook, khyathic\}@allenai.org} \quad \texttt{jmhessel@gmail.com}}

\begin{document}
\maketitle
\begin{abstract}
\input{00_abstract-v2}
\end{abstract}

\section{Introduction}
\label{sec:intro}
\input{01_intro-v2}

\section{Modeling Uncertainty in Conversations}
\label{sec:methods}
\input{03_methods}
\section{Experiments}
\label{sec:results}
\input{04_results}
\section{Related Works}
\label{sec:related}
\input{02_related}
\section{Conclusion}
\label{sec:conclusion}
\input{06_conclusion}

\bibliography{anthology,custom}
\bibliographystyle{acl_natbib}
\clearpage
\appendix
\onecolumn
\section{Theory and Derivations}
\label{sec:theory_details}
\input{a2_theory}
\clearpage
\section{Additional Experimental Details}
\label{sec:results_details}
Additional experimental details are provided next. In general, anything we have missed here will be available in the code, which will be made public.
\input{a1_results}
\end{document}

%% file: 00_abstract-v2.tex
Effective interlocutors account for the uncertain goals, beliefs, and emotions of others. But even the best human conversationalist cannot perfectly anticipate the trajectory of a dialogue. %Then the question arises:
How well can language models represent inherent uncertainty in conversations?
We propose \methodname, an expansion of the long-standing ``conversation forecasting'' task: instead of just accuracy, evaluation is conducted with uncertainty-aware metrics, effectively enabling abstention on individual instances.
We study two ways in which language models potentially represent outcome uncertainty (internally, using scores and directly, using tokens) and propose fine-tuning strategies to improve calibration of both representations.
Experiments on eight difficult negotiation corpora
demonstrate that our proposed fine-tuning strategies (a traditional supervision strategy and an off-policy reinforcement learning strategy) can calibrate smaller open-source models to compete with pre-trained models 10x their size. 

%% file: 01_intro-v2.tex
Dialogue models are increasingly fluent, topical, and informative conversationalists, capable of predicting plausible next-utterances given a partial conversation. Yet, the capacity to generate a single, plausible utterance is not the same as modeling the \textit{uncertainty} about all possible next-utterances in a calibrated way -- that is, assigning an appropriate probability to potential conversation outcomes, reflective of the randomness we observe in the real world. For example, in negotiations, ``Sounds good!'' or ``No thanks'' may be equally fluent/topical/informative next-utterances, but one choice may be more likely if the \textit{goals}, \textit{beliefs}, and \textit{emotions} of the interlocutors are taken into account. 
While even the best conversationalists cannot perfectly predict the trajectory of a dialogue, humans often manage uncertainty about social cues appropriately \citep{druckman2008emotions}, and demonstrate ability to both anticipate and affect the \textit{likelihood} of future conversation outcomes \citep{ho2022planning}. Meanwhile, it is not yet clear if language models posses even the simplest of these capabilities: \textit{anticipation of outcome certainty}.
\begin{figure}
    \centering
\includegraphics[width=\columnwidth]{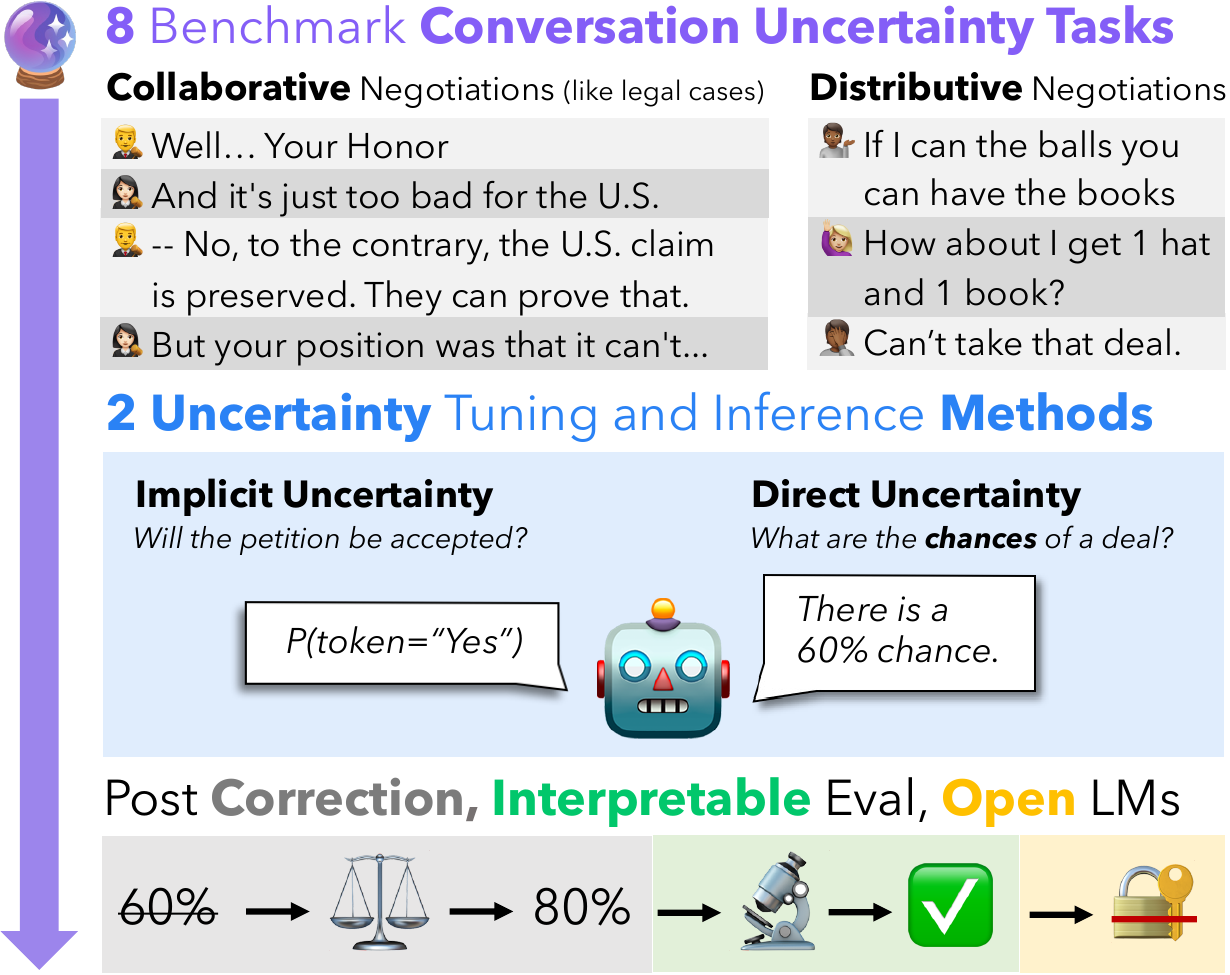}
    \caption{\methodname \ tests the ability of language models to represent uncertainty about future conversation outcomes. To meet this task, we tune models to express uncertainty directly in their output tokens or implicitly in their score distributions. We also provide additional strategies to correct uncertainty at inference-time. We propose tasks across 8 datasets, experimenting with GPT-4, Llama-2, and Zephyr-style models to release our best performing models publicly.}
    \label{fig:enter-label}
\end{figure}

To study this, we expand %\hyunwoo{"expand" seems to fit better?} 
the long-standing ``conversation forecasting'' task \citep{sokolova-etal-2008-telling, zhang-etal-2018-conversations}. While the usual goal is to predict the outcome of an unfolding dialogue, we instead account for how well language models represent \textit{uncertainty} about outcomes by measuring performance with calibration metrics. In effect, these calibration metrics allow models to abstain from predicting on instances when they estimate high uncertainty. Potential applications of models performant in this setting include: % we propose for this suite of tasks
% provide outcome-probabilities in diverse scenarios with foreseeable  in
improved tools for studying the effects of strategy and social structure in negotiations \citep{curhan2007thin}, intervening to improve human and machine conversations \citep{lewis-etal-2017-deal, zhou-etal-2019-dynamic, schluger2022proactive, argyle2023leveraging}, or assessing trust/heterogeneity in a data source via metrics like entropy \citep{csaky-etal-2019-improving, kuhn2022semantic}.%That is, whether language models can forecast the \textit{likelihood} of future outcomes in a conversation.

Here, we focus on the case of negotiations; this type of conversation is not only particularly sensitive to social uncertainties, but also, outcomes are readily quantified \emph{post-hoc}. %due to their sensitivity to prospective uncertainties as well as their quantifiable outcomes.
We ask language models questions about the likelihood of \textit{deals}, \textit{decisions}, and \textit{emotional conflicts} in settings like \textit{marketplaces}, \textit{online forums}, and \textit{courtrooms}, totaling 8 tasks to test uncertainty modeling in negotiations. 
%\jack{we didn't collect these; can we claim they are new tasks?} 
Our contributions include:

%In forming this task and investigation of baselines, we provide a host of technical contributions:

\begin{enumerate}[leftmargin=*,topsep=0pt,itemsep=-1ex,partopsep=1ex,parsep=1ex]
    \item formalizing the conversation uncertainty modeling task, along with its metrics (\S~\ref{sec:methods_problem}); % , with careful attention to the pitfalls of considering probability calibration alone;
    \item introducing two methods for representing uncertainty about the outcome of conversations using language models (\S~\ref{sec:if_and_df_desc});
    \item and proposing fine-tuning (\S~\ref{sec:sft}, \S~\ref{sec:utune_df}) and inference-time strategies (\S~\ref{sec:post-process}) for improving these representations.
    % % \item proposes new inference-time techniques to correct uncertainty estimates using limited prior knowledge (data);
    % \item and finally, proposes interpretable evaluation based on statistical significance (\S~\ref{sec:interp_eval}) \jack{it feels weird to introduce metrics after our main experiment results}
\end{enumerate}
We call this task \methodname.\footnote{\textbf{For}ecas\textbf{t}ing \textbf{Un}c\textbf{e}rtainty in \textbf{Dial}ogue.} %with \textbf{L}anguage \textbf{M}odels.} 
Experiments (\S~\ref{sec:results}) show GPT-4 and other large models can anticipate outcome certainty well, improving over prior knowledge by up to 9\%. Moreover, results show the utility of our fine-tuning strategies: smaller (7B) models are tuned to outperform pre-trained, open-source models 10x their size. Indeed, metrics improve up to 11\% on the tuning datasets and up to 3\% out-of-distribution. In the most difficult settings (out-of-distribution with incorrect prior knowledge), our fine-tuned models still meet the performance of their larger counterparts.
Besides the performance of our model deliverables, experiments also communicate insight on the biases of pre-trained language models at this task, the ability of different models to make use of prior knowledge, the impact of model/data scale, and the generalization of different algorithmic strategies.
% All of which culminates to the release of a suite of (relatively) small, fine-tuned models, which we call \methodname,\footnote{\textbf{For}ecas\textbf{t}ing \textbf{Un}c\textbf{e}rtainty using \textbf{L}anguage \textbf{M}odels.} \jack{It feels a bit odd to me to call a suite of models FortuneLM} that rival models 10 times their size and nearly meet the performance of commercial-scale models. \jack{This result does appear in the abstract, but, it would be nice to give more concrete results like this throughout, e.g., how much do our methods improve vs. zero shot? What are the shortcomings of models (do they over/under-estimate consistently?). It seems like our intro/abstract could have a few more concrete pointers like this.} % Our experiments in \S~\ref{sec:results} and the Appendix also communicate a myriad of insights on the biases of language models at this task, the ability of different models to make use of prior knowledge about the forecasting task, the impact of model/data scale, and the generalization of different tuning techniques. All
Models, code, and data will be made open-source. % upon publication.

%% file: 03_methods.tex
\begin{comment} % I am not sure we need this sec --- JMH
We
%propose to
model uncertainty in conversations via probabilistic forecasts of outcomes, i.e., we assign probabilities to different outcomes throughout the development of a dialogue, taking new utterances into account to update our measure of uncertainty. Uncertain outcomes have forecast probability near uniform,
%\hyunwoo{Do we want to always make it into a binary result?}
and more certain outcomes will tend towards 0 or 1. This type of uncertainty measure has value in studying negotiation strategies or proposing interventions, and moreover, can be used to build aggregate measures of uncertainty like \textit{entropy} \citep{berger-etal-1996-maximum, frank2010uncertainty, pimentel2021homophony, arora-etal-2022-estimating}. Yet, this is all predicated on the quality of our forecasts, insofar as they represent reality.
We formalize this problem and its evaluation, next.
% later tying in our use of language models as a solution.
\end{comment}
\subsection{Problem, Notation, and Evaluation}
\label{sec:methods_problem}
% \paragraph{The Forecasting Task} 
Consider a natural language token set $\mathcal{T}$. %, which includes \jack{2?} special tokens to delimit speaker turns. 
We observe partial multi-party dialogues $D \in \mathcal{T}^*$ consisting of $K \sim \mathcal{U}\{2, L\}$  % \jack{I assume this should be L-1?} 
turns, with $L+1$ being the eventual (random) length of the full dialogue.
Speaker turns are delimited by special sequences of tokens; e.g., ``Speaker 4: ...''
% (random)
% \jack{is the actual length of the full dialogue random?} -> technically, yes (speakers can terminate the conversation at any point, 
% \jack{This is a somewhat unfamiliar way of describing a conversation. How many participants are there? Do we represent turns? Are the conversations split at the utterance level?}
These partial conversations
are unfinished, but have eventual outcome $O \in \mathcal{O} = \{0,1\}$.\footnote{We only consider binary outcomes, but are flexible in application of this formulation, e.g., we can ask multiple questions to handle more outcomes in a one-vs-all fashion.}
% A partial dialogue is a roll-out of $K$ turns. In \S~\ref{sec:results}, $K \sim \mathcal{U}\{2, L\}$ with $L$ the (random) length of the full dialogue.
% I am going to un-footnote this: \footnote{A token is a word part and a dialogue is uniquely represented by tokens using keywords for speaker delimitation. A partial dialogue is a roll-out of $K$ turns. In \S~\ref{sec:results}, $K \sim \mathcal{U}\{2, L\}$ with $L$ the (random) length of the full dialogue.}
% For simplicity, we do not consider other multi-modal context. }
Nature picks a \textbf{\textit{conversation distribution}} 
%\hyunwoo{conversation distribution?} 
$\mathbb{D}$ over $\mathcal{T}^* \times \mathcal{O}$ which governs our supervised observations: $(D, O) \sim \mathbb{D}$. A \textit{\textbf{forecaster}} $f$ maps $D \mapsto \hat{P} \in [0,1]$ where $\hat{P}$ estimates the probability $O = 1$.
\paragraph{Evaluation with Proper Scores} 
%Commonly, forecasts are evaluated by their calibration. \jack{Commonly? Like: commonly in our paper? If so, we probably don't need to say commonly because it gives the impression that our new formulation isn't new, it's actually common practice to do this.} 
A \textit{\textbf{calibrated}} forecaster satisfies \citep{brocker2009reliability}:
% meeting with anthony on 1/10
% \jack{This equation has several factors that don't add up for me. How can we consider a distribution over all of $\mathcal{T}^*$ when a vast majority of those ``conversations" are jibberish sets of tokens? And does it matter that that set is infinitely large? And what is expectation of $O$ when it's not conditioned on the conversation? Do we even need this equation? }
% \jack{we can update according to discussion on 1/10, maybe some conditioning on D, or...}
\begin{equation}
\label{eqn:calibration}
\mathbf{E}[O \mid \hat{P} = p] = p \quad \forall p \in \{f(x) \mid x \in \mathcal{T}^*\},
\end{equation}
which intuitively means if we consider all conversations assigned $p$ by the forecaster, the mean occurrence of the outcome should also be $p$.
While commonly used to asses the verity of general probability estimates \citep{guo2017calibration}
the constraint in Eq.~\eqref{eqn:calibration} is often too broad
%, selecting a group of forecasters that don't all match the true conditional probability, as in Eq.~\eqref{eqn:exactmatch}.
because calibration, by itself, fails to measure the \textit{variance} of a forecast \citep{ovadia2019can}.
For example, the constant forecast $\hat{P} = \mathbf{E}[O]$ is calibrated, but rarely captures the
\textit{true} outcome probability (conditioned on the conversation). The issue of variance is \textit{especially} important in our setting, where social and temporal uncertainties make anticipation difficult; i.e., the basic, indiscernible prediction $\mathbf{E}[O]$ may be competitive. To accommodate calibration \textit{and} variance, we consider the constraint
\begin{equation}
\label{eqn:exactmatch}
\hat{P} = P \overset{\text{def}}{=} \mathbf{E}[O \mid D]
\end{equation}
% which is more specific and correct, requiring us to model the \textit{true} uncertainty about an outcome. \citet{gneiting2007probabilistic} suggest one way to achieve Eq.~\eqref{eqn:exactmatch} which they observe as ``maximizing variance % \hyunwoo{why do we maximize variance..?} -> this is mentioned above 44 in latex, 122 in text. I changed the wording in the presentation of the constraint just above to indicate we are still talking about variance when suggesting that. Let me know if it's not clear still! - Anthony
% \textit{subject to} calibration.'' 
One way to achieve this is by optimizing a \textit{\textbf{scoring function}} 
$\mathsf{s} : [0,1] \times \mathcal{O} \to \mathbb{R}_{\geq 0}$:
\begin{equation}
\label{eqn:scoring}
\min\nolimits_f \mathbf{E}[\mathsf{s}(\hat{P}, O)].
\end{equation}
If the scoring function is \textit{\textbf{strictly proper}},\footnote{Strict propriety requires that $\mathbf{E}[\mathsf{s}(P, O)] \leq \mathbf{E}[\mathsf{s}(\hat{P}, O)]$ for all $\hat{P}$ with equality if and only if $\hat{P} = P$. } Eq.~\eqref{eqn:exactmatch} is satisfied by the minimizer of \eqref{eqn:scoring}, so solving \eqref{eqn:scoring} recovers the true uncertainty %\jack{(implicit conditioning:) Doesn't the true outcome probability depend on significantly more than just the tokens of the conversation?}
%\jack{is $\mathbf{E}[O|D] $ the same as $P(O|D)$?}
as desired. Moreover, Eq.~\eqref{eqn:scoring}, indeed, optimizes variance and calibration equally, among other nice properties for ranking suboptimal forecasts \citep{brocker2009reliability}. 
\paragraph{Tangible Scores}
We use proper scores only, such as the \textbf{Brier Score} (\textbf{BS}; \citealp{brier1950verification}), which is the mean squared error between forecast probabilities and true outcomes.
While the use of proper scores is important (see previous), they do present some caveats: (1) they lack interpretable units and (2) for fixed tasks, they often vary on a small scale.\footnote{This is perhaps the reason for common \textit{improper} choices like calibration error, that do not account for forecast variance.} 
%; changes in Brier score do not have an interpretable unit. 
To resolve these issues, we sometimes focus evaluation on the \textbf{Brier skill score}: 
\begin{equation}
\label{eqn:skill-score}
    \textbf{BSS} = 1 - \textbf{BS} / \textbf{BS}_\texttt{ref}
\end{equation}
where \textbf{BS} is the Brier score of the forecaster we are evaluating and $\textbf{BS}_\texttt{ref}$ is the Brier score of some reference forecaster. One way to interpret the skill score is \textit{the percent improvement} of the forecaster compared to the reference. A simple reference, first proposed by \citet{brier1950verification}, is the constant prediction $\mathbf{E}[O]$, in which case $\mathbf{BS}_\texttt{ref}$ happens to be the \textit{variance} of the outcome. Here, we may interpret the skill score as \textit{the percent of variance} in outcome that is explained by the forecaster, like an $R^2$-value. On the other hand, $\mathbf{E}[O]$ can also be viewed as \textit{prior knowledge}, obtainable before observing $D$, implying skill conveys improvement over our prior knowledge. As desired, skill score tends to vary more than Brier score, while also having an interpretable unit (percentage).
\subsection{Language Models as General Forecasters}
\label{sec:if_and_df_desc}
An (auto-regressive) language model $\mathtt{LM}_\theta$ is a function parameterized by $\theta \in \mathbb{R}^d$ that returns a distribution over the next token $t \in \mathcal{T}$ conditional to any \textit{prefix} $x \in \mathcal{T}^*$. 
% (LM) is a probability kernel returning a distribution over the next token $t \in \mathcal{T}$ conditional to any prefix $x \in \mathcal{T}^*$; i.e., a model $\mathtt{LM}_\theta$ with $\theta \in \mathbb{R}^d$ satisfies:
% \begin{equation}
% \mathtt{LM}_\theta : \mathcal{T}^* \to [0,1]^{\lvert \mathcal{T} \rvert }, \ \forall x \ \mathtt{LM}_\theta(x) \in \Delta(\mathcal{T})
% \end{equation}
% where $\Delta$ is the set of distributions over $\mathcal{T}$. % \jack{Sure, but... this feels a bit convoluted, no? ACL audience will probably be familiar with LMs, and I am not sure this notation buys us anything.}
%\footnote{Measurability constraints are omitted \citep{kallenberg1997foundations}, which are unimportant for typical (discrete) text applications.} 
We write $T \sim \mathtt{LM}_\theta(x)$ for a \textit{single token sample} and $T \sim \mathtt{LM}^*_\theta(x)$ for the \textit{iterated sampling process}, 
%\jack{sometimes called teacher forcing} 
wherein we append a sampled token to $x$ and re-sample until a stopping condition.
% \hyunwoo{Not sure why you refer this as "latter case".}
%In practice, these models are optimized to reflect human text distributions on the web \citep{openai 1} or conversational text distributions that elicit high human satisfaction \citep{openai 2}. 
\begin{comment} % I think we can probably cut this text --- motivating LLMs feels somewhat out of place here. --- jmh
Language models are of interest for their unique ability to generalize to new data and tasks after large scale pre-training \citep{radford2019language, brown2020language} and tuning to human preferences \citep{ouyang2022training}.
This is done primarily by coaxing of their conditioning text, or \textit{prompting}.
\end{comment}
We define a \textbf{prompt} $\Phi$ as a function $\Phi: \mathcal{T}^* \to \mathcal{T}^*$ such that for any input $x$, it holds that $x$ 
%\jack{where does this $x$ come from?} 
is substring of $\Phi(x)$. So, $\Phi$ takes an input text $x$ and modifies it to a new text $\Phi(x)$, which contains the original text and (usually) adds important meta-information for solving the task; e.g., goal descriptors, expected output, and other context. 
% \jack{OK, but again, this does feel a bit convoluted for the notation, unless it buys us something later.}
% Jack, I think we talked about this if I remember and its all good now? haha - Anthony
We consider two types of prompts, which can turn a language model into a probability forecaster:
\begin{enumerate}[leftmargin=*]
    \item \textbf{Implicit Forecasts} (\texttt{IF}): The prompt $\Phi_\mathcal{O}$
    %, dependent on the outcome represented in $\mathcal{O}$, 
    poses the question ``Given the partial dialogue $D$, will the outcome represented by $\mathcal{O}$ occur?'' Then, the language model forecasts as
    \begin{equation}
    \hat{P}_\texttt{IF} = \mathbf{P}\{T = \texttt{yes}\}; \ T \sim \mathtt{LM}_\theta \circ \Phi_\mathcal{O} \circ D
    \end{equation}
    % \jack{is $\circ$ concatenation? Did we try putting it after D?}
    where $\texttt{yes} \in \mathcal{T}$ is an affirmation token
    %\jack{Did we try normalizing by the probability of No?}
    %\footnote{Candidate sets can also be used to infer a ``normalized'' %\jack{this should be ``normalized" I think, but I will let the more knowledgable folks change.} 
    % probability \citep{jiang2021can}. We discuss this in \S~\ref{sec:if_comp}.} 
    % Removing footnote since it's now redundant to two paragraphs later - Anthony
    and $\circ$ is function composition.
    % So, the model's internal representation of confidence in its answer is used as an uncertainty measure.
    \item \textbf{Direct Forecasts} (\texttt{DF}): The prompt $\Phi_\mathcal{O}$ poses the modified question ``Given the partial dialogue $D$, \textit{what is the probability} the outcome represented by $\mathcal{O}$ will occur?'' Then, the model forecasts as:
    \begin{equation}
    \label{eqn:direct_forecast}
    \hat{P}_\texttt{DF} = \mathsf{p} \circ T; \quad T \sim \mathtt{LM}^*_\theta \circ \Phi_\mathcal{O} \circ D
    \end{equation}
    where $\mathsf{p} : \mathcal{T}^* \to [0,1]$ is a parser that extracts a ``probability estimate'' from sample $T$; 
    % \jack{Does it output in natural language?}
    i.e., the model answers directly in natural language. 
\end{enumerate}
More details of exact prompts are in \S~\ref{sec:results}. Abstractly, both prompts provide a language description of the uncertainty modeling task, but make different assumptions.
\begin{comment} % JMH: I think we can cut this.
    about the best way to add to solve these tasks. In the former (\texttt{IF}), we assume that the probability of an affirmation in language roughly reflects the likelihood of affirmation in reality. In the latter (\texttt{DF}), we assume that direct expressions of probability in language (sampled according to our model) are better reflections of reality. We also consider how these properties can be fine-tuned, next, which we expect to be valuable for smaller language models.
\end{comment}
% As neither of these is likely to be the case by default, we study algorithms for fine-tuning these behaviors, next.
% \hyunwoo{I think this sentence gives the impression of we're doing something unusual or odd with an LM. Perhaps remove this?}
\subsection{Uncertainty Tuning of Implicit Forecasts}
\label{sec:sft}
We consider a language model with pre-trained parameters $\theta_\text{init}$, e.g., pre-tuned to follow instructions \citep{ouyang2022training}. The model computes a score vector $Z | \Phi(D) \in \mathbb{R}^{\lvert \mathcal{T} \rvert}$ and uses $Z$ to forecast:
\begin{equation}
\label{eqn:softmax_proba}
\hat{P}_\texttt{IF} = \mathbf{P}\{T = \texttt{yes}\} = \frac{\exp(Z_{\texttt{yes}} / \tau)}{\sum_{t\in\mathcal{T}} \exp(Z_{t} / \tau)}
\end{equation}
where temperature $\tau$ is a fixed hyper-parameter. A fine-tuning objective can then be written:
\begin{equation}
\label{eqn:fine-tuning}
\max_{\theta \ : \ \theta_\text{init} \rightarrow \theta} \mathbf{E}[O \ln \hat{P}_\texttt{IF}  + \overline{O} \ln \mathbf{P}\{T = \texttt{no}\}]
\end{equation}
where $\mathtt{no}$ is a dis-affirmation token. In effect, Eq.~\eqref{eqn:fine-tuning} translates the objective in Eq.~\eqref{eqn:scoring} to a fine-tuning objective by picking $\mathsf{s}$ to be the negative log score (a proper score, essentially equivalent to standard cross-entropy). \citet{jiang2021can} also consider calibration of pre-trained language models by direct supervision (as above), but focus on ``factual'' question-answering tasks where answers are more clearly right/wrong and the inherent social/temporal uncertainties of conversation are absent. In addition to a difference of setting, our proposal also differs from \citet{jiang2021can} because we retain the language model's \textit{whole} token distribution during inference, instead of only a candidate set. In \S~\ref{sec:if_comp}, we provide a first theoretical and empirical characterization of the impact of this choice when fine-tuning language models. Our main observation is these techniques are practically equivalent at inference-time with less than 1\% average difference in forecast (for our corpora). Thus, we advocate to retain the whole token distribution, since it is more easily coupled with other language modeling tasks; e.g., it doesn't require special machinery, like a loss with separate normalization.
% in Eq.~\eqref{eqn:softmax_proba}, which means our approach can be incorporated into a larger (e.g., instruction) tuning process.
% %\jack{I think even the binary version where only the yes/no logits are optimized could also be incorporated into a larger tuning process} --- yes, but we would need special machinery for this, e.g., masking
% \jack{I don't quite understand the connection to instruction tuning here. Is the implication that this should be an additional pretraining task if we are fine-tuning a model on dialogues? If so, we might want to be explicit about that and front it, because it feels like a bit of a random aside when mentioned here.} We theoretically and empirically characterize the impact of this design choice on the forecast, in \S~\ref{sec:if_comp}. \jack{What is ``this" choice? 
%  softmax over the whole vocab vs softmax over the two tokens?} \jack{If we fine-tune, we should definitely be explicit about it here --- this is basically just the supervised version of the task, either with logistic regression, or with logistic regression, right?}
% % \hyunwoo{We can probably remove the following if we need more space for other discussions.}
% %and the notation $\theta_\text{init} \rightarrow \theta$ denotes the dependence of the fine-tuning process on the initial pre-trained parameters.\footnote{When optimizing Eq.~\eqref{eqn:fine-tuning} by stochastic gradient ascent (fixed rate, finite time, bounded gradients), the initial parameters provably restrict solutions to a bounded set around $\theta_\text{init}$.}
\begin{figure*}
    \centering
    \includegraphics[width=\textwidth]{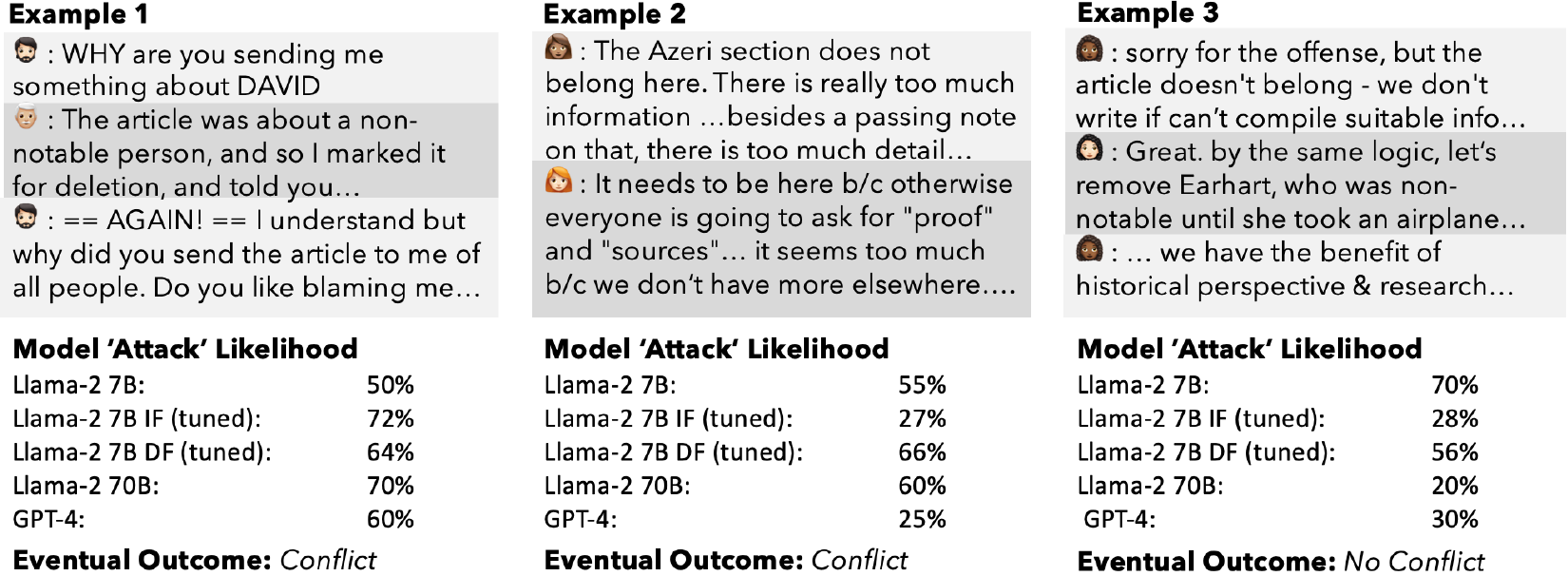}
    \caption{Examples of model forecasts for the eventual occurrence of a \textit{personal attack}. Models receive priors from data (\S~\ref{sec:post-process}) without any forecast scaling. Tuning (\S~\ref{sec:sft}, \S~\ref{sec:utune_df}) improves 7B parameter models and GPT-4 shows bias against conflict, compared to other models (\S~\ref{sec:results}). The nuances that lead to conflicts are not necessarily obvious.}
    \label{fig:examples}
\end{figure*}
\paragraph{Sampling Distribution}
%For brevity, Eq.~\eqref{eqn:fine-tuning} under specifies the sampling of $T$. \jack{This sentence feels a bit weird... it's a bit deflating because it almost comes across as if we are lying to the reader earlier... We could probably just skip that sentence?} 
In practice, we consider several negotiation datasets, defining distributions $\mathbb{D}_1 \ldots \mathbb{D}_\ell$ and prompts $\Phi_1 \ldots \Phi_\ell$. At test-time, if dataset diversity is sufficient, we expect the forecasts will generalize to new, \textit{possibly unseen}, environments $\mathbb{D}_{\ell+1}$ and prompts $\Phi_{\ell+1}$. 
% At test-time, we cross our fingers and hope the forecasts will generalize to new, \textit{possibly unseen}, negotiation environments $\mathbb{D}_{k+1}$ and prompts $\Phi_{k+1}$. 
% \hyunwoo{It's true, but doesn't this sound a bit too casual? lol}
%\jack{Do we need this formal terminology? This is just train/test mismatch, no?} 
Formally, this setup is called \textit{domain generalization} \citep{blanchard2011generalizing, muandet2013domain} and, while many approaches to this problem exist, simply training on the balanced aggregate of all available domains often performs best in practice \citep{gulrajani2020search}; we take this approach in \S~\ref{sec:results}.
% and continue to omit any dependence on domain indices, since it does not effect our formal discussion.
\subsection{Uncertainty Tuning of Direct Forecasts}
\label{sec:utune_df}
% One important observation is that language models also have the ability to \textit{directly} forecast a probability in their output samples; i.e., via tokens \citep{tian2023just}. Natural language tokens can be used to explicitly write probabilities, like ``72\%'' or ``0.72'', and more abstractly, express confidence by the words used \citep{mielke2022reducing}; e.g., ``I am \textit{certain} the answer is yes.'' To remain general, we assume existence of a \textit{parser} $\mathsf{p} : \mathcal{T}^* \to [0,1]$, which (semantically) extracts the LMs intended probability assignment from the sampled output tokens. Granted  $\mathsf{p}$, the LM forecast is defined: 
% \begin{equation}
% \label{eqn:direct_forecast}
%     \hat{P}_\texttt{MDP} = \mathsf{p} \circ T; \quad T \sim \mathtt{LM}^*_\theta \circ \Phi_\mathbb{D} \circ D.
% \end{equation}
% Where $ \Phi_\mathbb{D}$ should now question the likelihood of an outcome, rather than its occurrence.
% \hyunwoo{Section 3.3 and 3.4 titles don't seem to get along well. Probably would be better to imply that we're tuning the direct forecasting too?}
% \jack{Something about the motivation for expressing uncertianty in language vs. the yes or no tokens: motivation = maybe the performance is higher if we allow the model to verbalize beyond "yes" "no", but maybe there's somethign more fun to say :-) }
% In factual question-answering tasks, pre-trained language models often express better calibrated signals of certainty about factual correctness direc, compared to implicit scores \citep{tian2023just}.
Current pre-training strategies may prime models to express uncertainty best directly, via their output tokens; e.g. this is observed when models express uncertainty about factual correctness in question-answering \citep{tian2023just}. Ideally, despite the different setting, fine-tuning can preserve and capitalize on this predisposition. One challenge is that direct forecasts make Eq.~\eqref{eqn:scoring} non-differentiable, due to the parser. So, we formulate direct forecast tuning as a Markov Decision Process. We use reward $R = -\mathsf{s}(\mathsf{p} \circ T, O)$ % \jack{can we have an aside here about what this reward means in practice? This is a supervised reward, right, i.e., it requires O?} 
with $T \sim \mathtt{LM}^*_\theta \circ \Phi \circ D$ and set $\mathsf{s}$ to the log score (see Eq.~\ref{eqn:fine-tuning}). In effect, the reward is the negative score of our forecaster. Then, the usual objective $J(\theta)$ of this Markov Decision Process is:
%\hyunwoo{I think we didn't give a precise definition of $-\mathsf{s}()$ anywhere. I think I saw it somewhere before but looks like it got removed..?}
\begin{equation}
\begin{split}
\max_{\theta \ : \ \theta_\text{init} \rightarrow \theta} & \mathbf{E}[R] = -\min_{\theta \ : \ \theta_\text{init} \rightarrow \theta} \mathbf{E}[\mathsf{s}(\hat{P}_\texttt{DF}, O)].
\end{split}
\end{equation}
That is, we recover the original forecasting objective. While significant machinery has been developed for reward optimization (see \citealp{sutton2018reinforcement}) we apply policy gradient.
% This is an important observation.  In particular, framing forecasting as an MDP opens up this problem to the wide array of reinforcement learning algorithms designed to find optimal policies ($\theta$) for MDPs. For example, .
\subsubsection{Policy Optimization}
\label{sec:df-ft}
We focus on gradient-based policy optimization techniques, like REINFORCE \citep{williams1992simple} and PPO \citep{schulman2017proximal}. 
%\jack{why not DPO or PPO?} 
In particular, we derive an \textit{off-policy} version of the policy-gradient theorem, specific to our forecasting task, which uses Monte Carlo samples to produce unbiased estimates of the gradient-updates for our optimization problem. The \textit{off-policy} aspect is an important one. It means we can iteratively sample \textit{any} policy (distribution) over our token space $\mathcal{T}$, and use these demonstrations to learn $\theta$. Thus, while tuning, we can prioritize \textit{exploration} vs. \textit{exploitation} however we like, which can be an important factor for acting optimally in very general environments \citep{jiang2023on}, as is desired by our framework.
\paragraph{Off-Policy Policy Gradient}
For any random variable $X$, define $\mu_X$ as the mass function of $X$. Then, for any reference model $\mathtt{Ref} : \mathcal{T}^* \to \Delta(\mathcal{T})$:
\begin{equation}
\label{eqn:policy-opt}
\begin{split}
    \nabla_\theta \mathbf{E} [R] = \mathbf{E} \Big [ & \mathsf{s}_{\tilde{T}} \cdot \tfrac{\mu_T(\tilde{T})}{\mu_{\tilde{T}}(\tilde{T})}\cdot \nabla_\theta \log \mu_T(\tilde{T}) \Big ] \\
    \text{where} \quad & T \sim \mathtt{LM}^*_\theta \circ \Phi \circ D, \\
    & \tilde{T} \sim \mathtt{Ref}^* \circ \Phi \circ D, \\
    & \text{and} \ \mathsf{s}_{\tilde{T}} = -\mathsf{s}(\mathsf{p} \circ \tilde{T}, O).
\end{split}
\end{equation}
We derive this in \S~\ref{sec:pgt_pf}.
% As noted, the interpretation of Eq.~\eqref{eqn:policy-opt} is that we can iteratively sample any distribution over $\mathcal{T}$, compute the scoring function for this sample, and use this sample/score to optimize the reward of our language model $\mathtt{LM}_\theta$ on the forecasting task.
% \hyunwoo{Not sure whether this re-iteration for off-policy optimization is really necessary. Probably we can say something different, for example, it's possible to use (other) LMs with different prompts?}
While other off-policy policy gradient techniques exist \citep{degris2012off, imani2018off, kallus2020statistically}, the specifics of our problem allow us to make simplifying assumptions and yield a ``simpler'' and unbiased estimate of $\nabla_\theta \mathbf{E}[R]$ as the above.

% As a computational aside, it can be the case that the ratio of mass functions $u_T / u_{\tilde{T}}$ can be too large/small causing gradient explosion/vanishing.
% As a solution, we borrow the popular clipping strategy from Proximal Policy Optimization \citep{schulman2017proximal}; i.e., for $\epsilon \in [0,1]$, we use the update:
As a computational note, there may be instances where the ratio of mass functions $u_T / u_{\tilde{T}}$ becomes excessively large or small, leading to issues of gradient explosion or vanishing.
To address this, we adopt a widely-used clipping strategy from Proximal Policy Optimization \citep{schulman2017proximal}. Specifically, for $\epsilon \in [0,1]$, the update is:
\begin{equation}
\begin{split}
    & \mathsf{s}_{\tilde{T}} \cdot \omega \cdot \nabla_\theta \log \mu_T(\tilde{T}) \quad \text{where} \\
     & \omega = \min\Big \{ \max \Big \{\tfrac{\mu_T(\tilde{T})}{\mu_{\tilde{T}}(\tilde{T})}, 1-\epsilon \Big \}, 1+\epsilon \Big \}.
\end{split}
\end{equation}
Besides the computational benefits, this has motivations related to on-policy trust-region optimization \citep{schulman2015trust}, when $\texttt{Ref} = \texttt{LM}_\theta$.
% We leave exploration of this to future work.

\paragraph{Off-Policies} We consider three main choices for the off-policy reference $\texttt{Ref}$ in this work:
\begin{enumerate}[nolistsep]
    \item \textbf{The Explorer} \explore \ takes random actions in the token space $\mathcal{T}^*$, restricted only in the sense that it must output a description of the uncertainty (i.e., a probability). 
    %For example, practically speaking, this can be implemented by conducting a series of random coin flips, and reporting the outcome in textual format. 
    Exploration can benefit generalization \citep{jiang2023on}, since it exposes the agent to more diverse state-action pairs at train-time. 
    
    \item \textbf{The Exploiter} \exploit \ takes the actions that are optimal based on experience. For example, the tuned implicit forecasts $\hat{P}_\texttt{IF}$ can be mapped to a deterministic policy, and moreover, a near optimal one -- the log score in Eq.~\eqref{eqn:fine-tuning} is, indeed, a proper scoring function. While optimal on the training data, this may not be true for new conversation domains or outcomes.

    \item \textbf{The Quantizer} \interp \ takes actions learned from ground truth data by binning. It is inspired by \citet{lin2022teaching}, who calibrate model uncertainty to factual correctness by averaging labels of correctness for different ``bins'' of sub-tasks to estimate per task probabilities. The model is trained in a supervised manner to predict these estimates directly. Lacking clear ``sub-tasks'', we propose a (new) more general strategy, using clustering to bin our data. In \S~\ref{sec:results}, we also ablate our use of RL to optimize, instead of routine supervision.% \hyunwoo{Didn't this work kind of "quantized" the level of difficulty? Not sure whether it's accurate to say they interpolated. Maybe it would be better to just say "inspired by \citet{lin2022teaching}"? We'll probably need more details in the appendix too.}
    % In a loose sense, this simple approach provides a middle-ground between the two extremes. It is less sophisticated (and prone to over-fitting) than the pure exploitation strategy, while being slightly more sophisticated than taking random actions 
\end{enumerate}
Greater detail on these policies, including precise definitions and implementation are in \S~\ref{sec:pgt_detail}.
\subsection{Post-Hoc, Inference-Time Corrections}
\label{sec:post-process}
\paragraph{A New View on Old Tricks}
If validation data is available, temperature ($\tau$) scaling \cite{guo2017calibration} is common to correct the scale of implicit forecasts \cite{jiang2021can, kadavath2022language}. 
% \hyunwoo{Since equation 7 is used for implicit forecasting and this section is for direct forecasting, this can be a bit confusing.}
Yet, this helpful technique is not well-studied for direct forecasts 
% \jack{because of the discrete nature of the space/parser/some connection back to why its hard for that case}
because these are parsed from a discrete token sequence (they have no underlying latent scores to scale).
To address this, we propose a \textit{unified} correction strategy that is well-suited for both implicit and direct forecasts. We suggest estimation of the underlying latent scores to allow \textbf{post-hoc scaling} \scale \ for any forecast style:
\begin{equation}
\label{eqn:post-process}
\begin{split}
    \hat{Z}_\texttt{yes} & \leftarrow \log \hat{P} / (1 - \hat{P}) \\
    \tilde{Z}_\texttt{yes} & \leftarrow \hat{Z}_\texttt{yes} / \tau - \beta \\
    \hat{P}_\texttt{new} & \leftarrow 1 / (1 + \exp(-\tilde{Z}_\texttt{yes}))
\end{split}
\end{equation}
where $\tau$ relates to temperature as before and $\beta$ is a bias correction term. By estimating the latent score as above, we can effectively ``simulate'' traditional ($\tau$) scaling in such a way that it works for \textit{both} implicit and direct forecasts. We argue this theoretically and compare our proposal to other correction techniques in \S~\ref{sec:comp_post_hoc}, finding it theoretically equivalent, practically equivalent, or better in general. Ultimately, we decide to use it as the primary correction in \S~\ref{sec:results}.
% \jack{maybe an aside in language about what this is: ``estimate the logit" --- anthony} %\jack{Our novelty is the introduction of $\beta$? Do we need to? why is temperature scaling not enough?} \jack{What is $PP$?}
% Note, we also incorporate a bias correction $\beta$ as suggested for SVM calibration by \citet{platt1999probabilistic}. We compare our proposal to other correction techniques in \S~\ref{sec:comp_post_hoc}, and ultimately decide to use it as the primary correction in \S~\ref{sec:results}, finding it theoretically equivalent, practically equivalent, or better in general.
\paragraph{A Bayesian View} Use of prior knowledge is considered an important qualifier for generalization in some theories of machine learning \citep{mcallester1998some}. % \jack{This  sentence feels a bit odd/general. Why not just cite Bayes? We could skip this sentence/cite.} -> this specific cite is about a ML theory paper that uses Bayesian techniques, I tried to make it more specific
Motivated by this, we explore the use of natural language priors as a type of inference-time correction; e.g., , ``On average, this type of conversation ends with \{x\} about \{y\}\%...'' We explore three main types of \textbf{priors}: an average outcome learned from \textit{data}, a \textit{low} estimate of the data average, and a \textit{high} estimate of the data average.
% \jack{This feels like a modification to the prompt, no? Why didn't we talk about it with the other prompts?}
\paragraph{Correction is Not Always Zero-shot} Quality corrections require data in new domains. Data (n $\approx$ 250) is generally used to learn $\tau,\beta$ in Eq.~\eqref{eqn:post-process}, so scaling is not zero-shot. For priors, data can also be used or, in practice, a ``guess'' can be made. To simulate a ``guessed'' prior, we use averages \textit{outside} a 95\% confidence interval of the data average (n=50). Intuitively, this means we would rarely estimate this average using data (across repeated experiments), which is akin to, or worse than, human guesswork.
%\footnote{It can also leak something closer to the true mean, but in any case, this is conceptually akin to a human ``guess'' as well.} 
\S~\ref{sec:results} considers \texttt{data priors} learned from \textit{data} and \texttt{bad priors} using a \textit{high} and \textit{low} simulated guess.\footnote{We also checked non-numeric priors, e.g., reminding all outcomes have non-zero likelihood; these were worse overall.} We consider only the second to be zero-shot, since it tests robustness to ``guesses'' we expect to rarely obtain via data.  

%% file: 04_results.tex
\begin{table*}[]
\centering\small
\begin{tabular}{lll|lllll}
\textbf{Dataset}  & \textbf{Situation}                        & \textbf{Outcome}            & \textbf{\# Speak} & \textbf{\# Turn} & \textbf{\# Char} & \textbf{Aff.} & \textbf{Distr.} \\\hline
\citeauthor{zhang-etal-2018-conversations}      & wikipedia editing & personal attack    & > 2        & 6.2   & 2.5K   & yes & no       \\
\citeauthor{he-etal-2018-decoupling}  & craigslist       & best deal for buyer    & = 2         & 9.8   & 720   & no & yes        \\
\citeauthor{chawla-etal-2021-casino}   & camp provisions         & both camps happy & = 2         & 11.4   & 1.2K   & yes & yes      \\
\citeauthor{chang-danescu-niculescu-mizil-2019-trouble}       & reddit                       & personal attack    & > 2        & 5.3   & 3.2K   & yes & no       \\
\citeauthor{wang-etal-2019-persuasion} & charity      & donation occurs    & = 2         & 20.6   & 2.2K   & yes & no       \\
\citeauthor{lewis-etal-2017-deal}     & item allocation         & deal occurs        & = 2         & 5.0   & 253   & no & yes        \\
\citeauthor{10.1145/3359308}   & wikipedia editing & article deleted    & > 2        & 8.6   & 2.2K    & no & no        \\
\citeauthor{Chang2020ConvoKitAT}$^\text{a}$   & courtroom         & petitioner wins    & > 2        & 218   & 55K   & no & no \\
\end{tabular}
% \\\footnotesize{.}
\caption{Forecasting tasks. We list setting, outcome of interest, number of speakers, average turn/character count, and whether the setting is distributive. We also note if affective reasoning (about emotions) is useful in forecasting. Data are grouped into 3 train-test splits (\textit{easy}, \textit{med.}, \textit{hard}) to simulate generalization difficulty (see Tables~\ref{tab:imbalance}, \ref{tab:splits}). $^\text{a}$See also \citet{danescu2012echoes} for courtroom data}
\label{tab:data}
\end{table*}
% NEW RESULTS TABLES
\begin{table*}[]
\centering
\begin{tabular}{lllllllll|ll}\toprule
             & \multicolumn{3}{l}{\textit{post-hoc scaling \ \scale}}            & \multicolumn{5}{l}{\textit{no scaling \ \scale}} & \multicolumn{2}{l}{\textit{combined}}                                    \\ \cmidrule(lr){2-4}\cmidrule(lr){5-9}\cmidrule(lr){10-11}
             & \cancel{\texttt{prior}}      & \multicolumn{2}{l}{\texttt{data prior}} & \cancel{\texttt{prior}}  & \multicolumn{2}{l}{\texttt{data prior}} & \multicolumn{2}{l}{\texttt{bad prior}} & \multicolumn{1}{l}{\texttt{neg}} & \multicolumn{1}{l}{\texttt{all}}\\
             \cmidrule(lr){2-2}\cmidrule(lr){3-4}\cmidrule(lr){5-5}\cmidrule(lr){6-7}\cmidrule(lr){8-9}\cmidrule(lr){10-10}\cmidrule(lr){11-11}
\textbf{model}        & \textbf{BS} $\downarrow$            & \textbf{BS} $\downarrow$             & \textbf{BSS} $\uparrow$           & \textbf{BS} $\downarrow$       & \textbf{BS} $\downarrow$            & \textbf{BSS} $\uparrow$            & \textbf{BS} $\downarrow$            & \textbf{BSS} $\uparrow$ & \textbf{BI} & \textbf{BI}           \\\midrule
\texttt{gpt-4}              & 21.1          & 20.7           & 8.5           & 23.8     & 21.8          & 3.8            & 22.7          & 9.0    & -11  & -1.1       \\
\llama \ \texttt{70B IF}    & 22.9         & 23.1           & -2.0          & 66.1     & 66.1          & -182          & 66.1          & -160     & 3.6 & -49.4    \\
\llama \ \texttt{70B DF}    & 22            & 22.1           & 2.3           & 25.0     & 22.0          & 2.7            & 23.6          & 5.0      & -5.6  & -1.3     \\\bottomrule          
\end{tabular}
\caption{Scores for large, pre-trained models with different access to prior knowledge. Use of post-hoc correction and data-dependent priors is not truly zero-shot. GPT-4 tests use only direct forecasting (\texttt{DF}), while Llama-2-chat \llama \ 70B uses either direct or implicit (\texttt{IF}). \textbf{BSS} = 0 corresponds to constantly forecasting the prior, e.g., for \texttt{data prior}, this is the corpus mean outcome. \texttt{DF} consistently improves over prior knowledge (\textbf{BSS} > 0).}
\label{tab:large-models}
\end{table*}
% FINE-TUNED MODELS
\begin{table*}[]
\centering
\begin{tabular}{llllllllll|ll}\toprule
             & \multicolumn{2}{l}{\textit{in-domain}}            & \multicolumn{2}{l}{\textit{pseudo OOD}} & \multicolumn{5}{l}{\textit{zero-shot OOD}}                                    & \multicolumn{2}{l}{\textit{combined}} \\ 
             \cmidrule(lr){2-3}\cmidrule(lr){4-5}\cmidrule(lr){6-10}\cmidrule(lr){11-12}
             & \multicolumn{2}{l}{\texttt{all}} &  \multicolumn{2}{l}{\texttt{all}} &  \multicolumn{1}{l}{\texttt{easy}} &  \multicolumn{1}{l}{\texttt{med}} &  \multicolumn{1}{l}{\texttt{hard}} & \multicolumn{2}{l}{\texttt{all}} & \multicolumn{1}{l}{\texttt{neg}} & \multicolumn{1}{l}{\texttt{all}} \\
             \cmidrule(lr){2-3}\cmidrule(lr){4-5}\cmidrule(lr){6-6}\cmidrule(lr){7-7}\cmidrule(lr){8-8} \cmidrule(lr){9-10}\cmidrule(lr){11-11}\cmidrule(lr){12-12}
\textbf{model}  & \textbf{BS} $\downarrow$ & \textbf{LSS} $\uparrow$ & \textbf{BS} & \textbf{LSS} & \textbf{LSS} & \textbf{LSS} & \textbf{LSS} & \textbf{BS} & \textbf{LSS} & \textbf{BI} & \textbf{BI}\\ \midrule
\llama \ \texttt{7B}                                             & \multicolumn{1}{l}{}         & \multicolumn{1}{l}{}         & 22.7                         & -3.2                         & -10.9                         & -7.5                         & 1.1                          & 24.3                         & -5.1                         & -0.4                    & -2.7                    \\
$\hookrightarrow$ \texttt{IF}                                                                & \cellcolor[HTML]{E5F5E0}21   & \cellcolor[HTML]{A1D99B}5.3  & \cellcolor[HTML]{E5F5E0}22.4 & \cellcolor[HTML]{E5F5E0}-1.4 & -12.4                         & \cellcolor[HTML]{A1D99B}2.2  & -3.9                         & \cellcolor[HTML]{E5F5E0}23.9 & \cellcolor[HTML]{E5F5E0}-3.7 & 0.3                     & 0                       \\
$\hookrightarrow$ \texttt{DF} \exploit                                        & 22.8                         & -3.2                         & 22.9                         & -3.7                         & -14.1                         & -8.2                         & -8.3                         & 25.3                         & -9.7                         & 0.9                     & 0.6                     \\
$\hookrightarrow$ \texttt{DF} \explore                                        & 22.9                         & -3.9                         & 22.9                         & -3.6                         & \cellcolor[HTML]{E5F5E0}-9.3  & -8.5                         & \cellcolor[HTML]{A1D99B}1.7  & \cellcolor[HTML]{E5F5E0}24.2 & \cellcolor[HTML]{E5F5E0}-4.9 & 0.8                     & -0.9                    \\
$\hookrightarrow$ \texttt{DF} \interp \ $\texttt{rl}$          & 22.9                         & -3.7                         & 22.8                         & -3.3                         & \cellcolor[HTML]{E5F5E0}-10.7 & \cellcolor[HTML]{E5F5E0}-5.9 & \cellcolor[HTML]{A1D99B}2.1  & \cellcolor[HTML]{E5F5E0}24   & \cellcolor[HTML]{E5F5E0}-4.5 & 3                       & -0.6                    \\
$\hookrightarrow$ \texttt{DF} \interp \ $\cancel{\texttt{rl}}$ & 22.9                         & -3.8                         & 22.9                         & -3.7                         & -11.9                         & \cellcolor[HTML]{E5F5E0}-6.9 & \cellcolor[HTML]{A1D99B}2.7  & \cellcolor[HTML]{E5F5E0}24.1 & \cellcolor[HTML]{E5F5E0}-4.9 & 3.4                     & -0.9                    \\
$\hookrightarrow$ \texttt{IF}$^{\times 4}$                                                   & \cellcolor[HTML]{E5F5E0}19.6 & \cellcolor[HTML]{A1D99B}11.4 & \cellcolor[HTML]{E5F5E0}21.4 & \cellcolor[HTML]{A1D99B}3.7  & \cellcolor[HTML]{A1D99B}8.9   & \cellcolor[HTML]{A1D99B}2.9  & -6.9                         & \cellcolor[HTML]{E5F5E0}23.3 & \cellcolor[HTML]{A1D99B}0.7  & -3.5                    & -2.4                    \\
$\hookrightarrow$ \texttt{DF}$^{\times 4}$ \interp                            & 22.8                         & -3.2                         & 22.8                         & -3.4                         & \cellcolor[HTML]{E5F5E0}-10.7 & \cellcolor[HTML]{E5F5E0}-3.7 & \cellcolor[HTML]{A1D99B}4.5  & \cellcolor[HTML]{E5F5E0}23.6 & \cellcolor[HTML]{E5F5E0}-2.3 & 3.7                     & -2.8                    \\\midrule
\zephyr \ \texttt{7B}                                            & \multicolumn{1}{l}{}         & \multicolumn{1}{l}{}         & 22.5                         & -1.2                         & -9.9                          & -9.6                         & -8.1                         & 25.5                         & -9.1                         & -10.6                   & 1.8                     \\
$\hookrightarrow$ \texttt{IF}                                                                & 22.5                         & -1.6                         & 22.8                         & -3.4                         & -26.3                         & -10.2                        & \cellcolor[HTML]{E5F5E0}-6.8 & 25.8                         & -13                          & 1.5                     & -9.1                    \\
$\hookrightarrow$ \texttt{DF} \interp \ $\texttt{rl}$          & 23                           & -4.2                         & 23                           & -4.2                         & -11.4                         & \cellcolor[HTML]{E5F5E0}-8.9 & \cellcolor[HTML]{E5F5E0}-3.8 & \cellcolor[HTML]{E5F5E0}24.9 & \cellcolor[HTML]{E5F5E0}-7.6 & 0.2                     & -6.8                    \\\midrule
\zephyr \ \texttt{1B}                                            & \multicolumn{1}{l}{}         & \multicolumn{1}{l}{}         & 22.8                         & -3.3                         & -11.2                         & -3                           & -2.6                         & 24.4                         & -4.9                         & 3.5                     & 6.5                     \\
$\hookrightarrow$ \texttt{IF}                                                                & \cellcolor[HTML]{E5F5E0}22.2 & \cellcolor[HTML]{E5F5E0}-0.5 & 22.8                         & \cellcolor[HTML]{E5F5E0}-3.2 & -19.4                         & -10.6                        & -3.2                         & 25.2                         & -10                          & -2.4                    & -5.5                    \\
$\hookrightarrow$ \texttt{DF} \interp \ $\texttt{rl}$          & 23                           & -4.2                         & 23                           & -4.2                         & -11.4                         & -8.9                         & -3.8                         & 24.9                         & -7.6                         & 1                       & -1.2                    \\ \bottomrule
\end{tabular}
\caption{Scores for uncertainty tuned Llama-2-chat \llama \ 7B, Zephyr \zephyr \ 7B, and ``tiny'' 1B Llama-2 trained in Zephyr style. \textit{in-domain} shows test data scores from within tuning distribution. \textit{pseudo OOD} and \textit{zero-shot OOD} show scores when test data is out of distribution (i.e., held out domains). Val. data is used for post-hoc scaling and priors, except under \textit{zero-shot OOD}, which doesn't correct and uses ``bad'' priors. For \llama \ 7B, we also tune on $\times4$ more data. Improvements are highlighted: Light green \colorbox[HTML]{E5F5E0}{cells} show improvement against corresponding pre-trained models (same setup, before uncertainty tuning), while \colorbox[HTML]{A1D99B}{darker} cells (additionally) improve over the larger \llama \ 70B.}
\label{tab:tuned-models}
\end{table*}
\paragraph{Data \& Splits} % To study generalization of our uncertainty models,
We consider 8 modeling tasks spanning both traditional (distributive) negotiations and collaborative negotiations \citep{chu-carroll-carberry-1995-response}; Table~\ref{tab:data} summarizes the corpora. Tasks have diverse situations and outcomes, span multi-party/dyadic settings, and are both short/long. %Different forecasts may also benefit from affective reasoning (i.e., about people's emotions).
To simulate varying degrees of distribution shift, we group these datasets into different train/test splits, categorized as \texttt{easy}, \texttt{medium}, or \texttt{hard}. 
To make the forecasting task more difficult, each of these three splits hold out full datasets for testing. % while those data not heldout are used for training. 
Conceptually, the splits are designed to create different degrees of train/test imbalance for important properties like the topic, the length of the conversation, the type of outcome, and the number of speakers; Tables~\ref{tab:imbalance}+\ref{tab:splits} and \S~\ref{sec:data_details} provide more detail on train/val/test splits. % how these splits simulate degrees of difficulty more precisely.
% \jack{Surely we can provide a pointer towards how this works?} 
%\S~\ref{sec:data_details} provides other dataset details; e.g., train/val/test splits \textit{within} each dataset, used to select hyper-parameters or check model performance within the tuning distribution (in-domain).
%\jack{But dont we hold out full datasets? Why would we hae train/val/test splits within dataset?} 
Sometimes, we restrict inference to data with \textit{affective conflict} as outcome (\texttt{neg}) like, a personal attack or unhappy speaker. Here, to compute some metrics, we swap the positive and negative classes as needed, e.g., the positive class becomes $1-O$ to study ``both camps unhappy'' instead of ``both camps happy.''
% \hyunwoo{What does this last sentence mean..? What is \texttt{neg}? I can't seem to parse it.}

\paragraph{Models \& Prompts} We use GPT-4 (0613, \citealp{openai2023gpt4}), Llama-2-chat \cite{touvron2023llama}, and Zephyr-$\beta$ \cite{tunstall2023zephyr}, which have all been pre-tuned for chat/instruction following (Zephyr is pre-tuned via distillation). We also use a ``tiny'' Llama-2 replicate trained in the style of Zephyr (Chat-v0.6, \citealp{tinyllama}). Open-source model sizes range from 1B to 70B parameters. We use \textit{pre-trained} to refer to these models before we tune uncertainty (\S~\ref{sec:sft}, \ref{sec:utune_df}). Concretely, the prompts the models receive have: \textit{situational context} specific to each dataset, like ``the speakers are defending their opinions on an issue''; \textit{priors}, as in \S~\ref{sec:post-process}; and the \textit{main question} that asks the model about the likelihood of outcome occurrence in the conversation, or just occurrence for implicit forecasts. Pre-trained models also receive system prompts to constrain output and clarify the task goals. We use QLoRA \cite{dettmers2023qlora} for uncertainty tuning. Other details on prompts/hyper-parameters are in \S~\ref{sec:model_details}.
% \paragraph{Priors \& Post-Processing} Situational context and the main question are always provided in prompts, but prior information (like averages) may be available, or not. In a truly zero-shot setting, we may not have data to provide a quality estimate of the average outcome. Still, we find this type of prior can be \textit{vital} to a model. We explore three main types of \textbf{priors}: an average outcome learned from \texttt{data}, a \texttt{low} estimate of the data average, and a \texttt{high} estimate of the data average. Only the latter 2 priors can be considered truly zero-shot, since they represent realistic \texttt{bad} guesses of the average.\footnote{\texttt{low} / \texttt{high} averages are outside a 95\% CI of the data average, assuming n=50. We also checked non-numeric priors, reminding that all outcomes have non-zero chance to occur; these were (statistically) worse than ``bad'' numeric priors.} 
% NOTE: group all ZS bad priors together, then compare ``bad'' priors ``data'' priors and ``no'' priors for all models. Using high/low for some and not for others is confusing
% We also conduct post-hoc correction of forecasts as in \S~\ref{sec:post-process}. Validation data is used to learn parameters, so correction is also \textit{not} zero-shot.
\paragraph{Metrics} We use the Brier Score (\textbf{BS}) and skill score (\textbf{BSS}) as discussed in \S~\ref{sec:methods_problem}, macro-averaged across datasets and prompts. \textbf{BSS} refers to the original skill score \citep{brier1950verification} where the reference model in the skill score is the constant (average) outcome probability. When a prior is provided, we substitute this prior for the data mean in the reference score to account for how priors can implicitly bias forecasts. Indeed, variance around an (incorrect) prior is always higher than the true variance, so the prior-adjusted \textbf{BSS} reports the percent of this larger variance explained by the forecaster. Besides \textbf{BSS}, we also suggest a \textit{new} skill score called the \textbf{Llama Skill Score} (\textbf{LSS}). \textbf{LSS} is identical to usual skill scores, i.e., Eq.~\eqref{eqn:skill-score}, but uses the Brier score of Llama-2-chat \llama \ 70B (direct forecasts, same experimental setup) as the reference score $\textbf{BS}_\texttt{ref}$. This quantifies how smaller fine-tuned models compare to this large model by \% improvement. Finally, we report \textit{statistical bias} (\textbf{BI}) to convey average over- or under-estimation of outcome probability (positive or negative values, respectively).
\subsection{Results}
\paragraph{Forecasting is Better with Correction} Table~\ref{tab:large-models} shows Brier and skill scores of pre-trained models without uncertainty tuning. We modulate priors and ablate post-hoc scaling. GPT-4 has lower (better) Brier scores when granted access to validation data to make corrections, e.g., via data priors or scaling. Scaling appears to have the greatest impact on scores as they are lowest, even when excluding the data prior. However, priors are still useful. In zero-shot settings (i.e., no scaling or data prior), GPT-4 performs better when it has access to ``bad'' guesses of prior probability, rather than no guess at all. Trends are similar for Llama-2-chat  70B's direct forecasts.
\paragraph{Problems with Pre-trained Implicit Forecasts}
For Llama-2-chat 70B, we have access to scores of every token, so we can compare implicit forecasting to direct forecasting. In Table~\ref{tab:large-models}, implicit forecasting is worse for this pre-trained model. Echoing \citet{kadavath2022language}, we find post-hoc correction is \textit{vital} to improve pre-trained implicit forecasts. Moreover, the degradation of uncorrected implicit forecasts is quite high, suggesting amplification of this effect in our unique (conversational) setting.
\paragraph{Have Data? Tuned Implicit Forecasts Are Best} Based on the previous results, we focus on uncertainty tuning with access to a prior (even a ``bad'' one) and compare uncertainty tuned models to direct forecasts of pre-trained versions. We consider an \textit{in-domain} setting first, wherein test data follows the tuning distribution, post-hoc correction is used, and priors are data-dependent. Here, Table~\ref{tab:tuned-models} shows tuned implicit forecasts can significantly improve over pre-trained Brier score, even compared to models 10x their size. For instance, a tuned Llama-2-chat 7B improves over the 70B model by about 11\% (or, 5\% with less data). With enough training data, out-of-distribution (OOD) scores are also about 4\% better than Llama-2-chat 70B scores (if both have access to data for correction).
\paragraph{Direct Tuning Generalizes Better (Sometimes)} Next, we consider a zero-shot OOD setting \textit{without} data for correction. Here, performance of tuned implicit forecasts is still good, for Llama-2-chat 7B. With enough data, tuning brings Llama-2-chat 7B implicit forecasts to the skill of its 70B counterpart (+0.7\%), but for Zephyr-style (distillation-tuned) models, degradation is significant compared to (even) pre-tuning scores. In contrast, for all 7B models and different levels of data access, direct forecasts show consistent improvement of scores after uncertainty tuning (see light green cells). For Llama-2-chat 7B, tuned direct forecasts also handle difficult (\texttt{hard}) distribution shift up to 4\% better than their 70B counterpart. As speculated earlier, direct forecast tuning may preserve some predispositions of pre-trained models to direct uncertainty signals; this may explain consistent generalization of these forecasts \textit{beyond} the tuning distribution.
\paragraph{Qualitative Comparison of \texttt{IF} and \texttt{DF}} Tuned implicit forecasts tend to have higher variance than tuned direct forecasts. Averaging over all tuning strategies in Table~\ref{tab:tuned-models}, implicit tuning of Llama-2-chat 7B leads to 2$\times$ the standard deviation in forecasts compared to direct tuning (about 11\% and 5\% SD, respectively). As noted in \S~\ref{sec:methods_problem}, all else equal, a higher variance is preferred by our metrics in order to capture the discernability of forecasts. Potentially, directly tuned models tend towards distribution collapse due to insufficient regularization in our RL objective \citep{korbak-etal-2022-rl}. Direct tuning methods which resolve this issue will be of interest in future work.   
% \paragraph{Impact of Data Scale} Generally, where possible, we keep data consistent across tuning (and pre-training).\footnote{Excluding the unknown pre-train scale of Zephyr 7B and GPT-4, all models saw 2T tokens before any fine-tuning.} For the best Llama-2 setups, we also tune uncertainty using 4 times more data. Adding data improves performance, so much so that all tuned Llama-2 7B forecasts meet/exceed the zero-shot performance of Llama-2 70B. In-domain, tuned Llama-2 7B implicit forecasts can even surpass GPT-4.
\paragraph{Impact of Scale}
Tuning of implicit forecasts allow in-domain scores of a 1B model to rival a model 70x it's size (-0.5\%). But, tuning methods show less improvement, or even degradation, when applied to the 1B model OOD. Possibly, and especially since we use QLoRA tuning, the reduction in trainable parameters is detrimental to these methods. This, and the observed benefits of increasing data (see previous discussion), suggest our tuning techniques may follow neural scaling laws \citep{kaplan2020scaling}, meaning generalization is strongly dependent on model, data, and compute scale.
\paragraph{Exploration \& Exploitation} Llama-2-chat 7B findings indicate pure exploitation \exploit \ is detrimental to generalization when compared with pre-trained models (see absence of green scores). On the other hand, exploration \explore \ tends to offer some improvement, beating the pre-trained scores on \texttt{easy} and \texttt{all} as well as the 70B scores on \texttt{hard}. The best tuned direct forecasts use quantization \interp, which is neither completely random nor optimal on the train set, offering a balance of exploration/exploitation.
\paragraph{Benefits of RL} For our best performing direct forecast tuning mechanism, we also ablate the role of using RL to tune; i.e., we use a traditional supervised update rule, similar\footnote{Ours is still more general, due to the clustering proposal.} to \citet{lin2022teaching}. We find improvements over pre-training are less consistent and worse overall. 
\paragraph{Human Preference Tuning May Induce Bias} We also postulate human preference tuning (RLHF; e.g., \citealp{ouyang2022training}) may bias pre-trained models to under-estimate negative affective conflicts, since these are presumably undesirable to human annotators. To study this, we report statistical bias (\textbf{BI}) of forecast probabilities in predicting negative (\texttt{neg}) emotional outcomes; i.e., personal attacks, unhappy interlocutors, or refusals to donate to charity. On average, direct forecasts from GPT-4 underestimate this probability by about 11\%, while direct forecasts from Llama-2-chat 70B underestimate this probability by about 6\%. For the same models, bias across \texttt{all} outcomes is not as staggering. Smaller language models do not necessarily exhibit this bias, but distillation-tuned models may learn similar biases from their larger teachers (see \zephyr \ 7B). Generally, uncertainty tuning does not appear to introduce as staggering bias against emotional conflict, but especially for distillation-tuned models, the overall bias can be elevated.

%% file: 02_related.tex
\paragraph{Negotiation \& Conversation Forecasting}
% \citep{nouri-traum-2014-initiative} model outcome from simple features, just hard predictions
Negotiation and dispute modeling has a long history \citep{lambert-carberry-1992-modeling, jamesonKNSWZ94, traum2008virtual, lascarides-asher-2008-agreement} 
% koit-oim-2000-dialogue, cadilhac-etal-2013-grounding
with early works hand-crafting models of interlocutor behavior by logical or discourse structures. Reinforcement learning in simulated environments offers improvement \citep{georgila2011reinforcement, efstathiou2014learning} 
% georgila-2013-reinforcement, papangelis-georgila-2015-reinforcement
with most recent advances modeling opponents' dialogue acts \citep{keizer-etal-2017-evaluating}, word choices \citep{he-etal-2018-decoupling}, and mental states \citep{yang-etal-2021-improving, chawla-etal-2022-opponent}. Instead of full simulation, we focus on efficient and interpretable outcome models \citep{sokolova-etal-2008-telling, nouri-traum-2014-initiative}. Outcome models, or forecasts, are also common in broader dialogue for proactive moderation of social media \citep{zhang-etal-2018-conversations, kementchedjhieva2021dynamic} 
% chang-danescu-niculescu-mizil-2019-trouble
as well as predicting task-success \citep{walker2000learning, reitter-moore-2007-predicting}, mental health codes \citep{cao-etal-2019-observing}, emotions \citep{wang-etal-2020-sentiment, matero2020autoregressive}, situated actions \citep{lei-etal-2020-likely}, and financials \citep{koval-etal-2023-forecasting}. Among these, ours is first to propose and evaluate probabilistic methodology, modeling \textit{dynamic uncertainty} for the first time. Our proposal is also \textit{uniquely general}, operating independent of setting or outcome of interest. Indeed, we evaluate on general negotiations, looking beyond distributive applications (zero-sum games in specific markets) to include common \textit{collaborative negotiations} \citep{chu-carroll-carberry-1995-response}, like planning, where parties share some goals, but conflicts arise from competing sub-goals or beliefs.
\paragraph{Language Models \& Uncertainty}
Modern large language models fine-tuned to human preferences \citep{ouyang2022training} are increasingly general, ``unsupervised'' multi-taskers. When queried on factual information, these models also represent uncertainty about their solutions with little to no supervision \citep{kadavath2022language}. 
%That is, a model-asserted fact is correct about $p\%$ of the time, whenever the model's representation of confidence is also $p$\%. 
Uncertainty (about correctness) has also been studied in smaller models without preference tuning \citep{desai-durrett-2020-calibration, jiang2021can, dan-roth-2021-effects-transformer} with many algorithms for improvement \citep{kong-etal-2020-calibrated, zhang-etal-2021-knowing, li-etal-2022-calibration}. Albeit similarly operationalized, modeling uncertainty about factual correctness is distinct from our focus in negotiations, which elicits modeling of social dynamics and mental states. Fewer works study uncertainty in social reasoning \citep{jiang2021can, hosseini-caragea-2022-calibrating, kumar-2022-answer}. These look at smaller models without instruction-tuning, and lack focus on the interactive, temporal aspects of conversations that cultivate an inherent uncertainty about future outcomes. Ours is also one of few works that study how models communicate uncertainty directly via output tokens \citep{mielke2022reducing, tian2023just}, and fewer that propose tuning algorithms for this \citep{lin2022teaching}.
% \textcolor{red}{Multilingual language models (old and new) tend to be poorly calibrated on low-resource languages \citep{ahuja-etal-2022-calibration, krause-etal-2023-confidently}. Be sure to include this in a discussion or limitations.}
% removed for space
% as well as increasing application areas like machine translation \citep{wang-etal-2020-inference} and document ranking \citep{penha-hauff-2021-calibration}
% maybe cite
% \citep{park-caragea-2022-calibration} also mixup, on small early models, OOD is more meaningful here same dataset as desai so mostly still fact-oriented. OOD does not mean different tasks, just diferent datasets; e.g., i believe they use [CLS] token prediction rather than prompting
% \citep{chen-etal-2023-close} explore calibration on smaller models (RoBerta and T5-base) finding that training and size doesn't improve calibration
% \citep{dhuliawala-etal-2022-calibration} more factual (machine reading systems)
% probably won't cite:
% \citep{baan-etal-2022-stop} when not to use calibration - semantically irrelevant for this work i think
% \citep{si-etal-2022-examining} fact based QA, calibration evaluation - make the observation that ECE is bad, but don't realize proper scoring functions exist

%% file: 06_conclusion.tex
We show language models represent uncertainty about conversational outcomes quite well, depending on their size, inference strategy, training strategy, and access to prior knowledge. We design a task to evaluate this ability and show:
\begin{itemize}[nolistsep, leftmargin=*]
    \item large (commercial-scale) models do this well, provided limited data to pick hyper-parameters;
    \item without data, these models still offer improvement over low quality priors, like human guesses;
    \item specialized fine-tuning can elevate small open-source models to beat models 10x their size;
    \item current pre-trained language models may be predisposed to representing uncertainty in their textual outputs, instead of their logit distributions;
    \item exploration at train-time can be more beneficial for generalization (compared to exploitation);
    \item and finally, pre-trained models may be biased against forecasting emotional conflict. 
\end{itemize}
This work (and task) presents a first step towards understanding how language models can anticipate the certainty of outcomes in interactive social situations. We make our code, models, and data open-source to promote continued research.
% show some large (commercial-scale) models can do this quite well, provided small amounts of data to pick hyper-parameters. Even without data, these models can still offer improvement over low quality priors. We suggest fine-tuning methods for smaller open-source models, showing these can be tuned to beat models 10x their size. Besides fine-tuning algorithms, our work contributes useful insights on myriad practical considerations in language modeling, like inference-time forecast corrections, biases of pre-trained models, and generalization of different forecast strategies in conversational settings.
\section*{Limitations} While we explore a wide array of datasets and experimental setups, the generality of our conclusions are limited to what's explored in this paper. Further study, e.g., replication study with different data, models, and settings, would provide evidence to confirm the generalization of our findings. One aspect of particular interest, is the application of these techniques to languages other than English. Indeed, there is evidence that the uncertainty representations of language models may experience performance degradation when applied to other languages \citep{ahuja-etal-2022-calibration, krause-etal-2023-confidently}, especially those which are low-resource. 

Additionally, while our task is motivated by a desire to probe a language model's ability to anticipate social (un)certainties in conversation, there is no clear way for us to separate causation and correlation in this task. We cannot claim the language model actually ``understands'' the causes of social (un)certainties, like interlocutor mental states, since instead, it may be the case that language models capitalize on ``superficial'' or ``spurious'' statistical correlations associated with an outcome \citep{ho2022planning}.

Finally, we quantify the quality of a forecast primarily through its improvement over prior knowledge. Effectively, this compares a forecaster with a performance lowerbound, demonstrating the forecaster is using information revealed through the dialogue to provide an improved prediction. In the future, a performance upperbound (such as human performance) would be useful to establish a ceiling for our goals. This is particularly important for proper scores, since it is exceedingly rare for forecasters to achieve a perfect value (e.g., 0 for the Brier score or 1 for the skill score).
\section*{Ethics Statement} The models we use and train may exhibit, or even amplify, any biases  contained in the training data, such as societal biases. Moreover, robustness to adversaries and natural token perturbations is not guaranteed. In critical applications, ethical and safety considerations, such as bias mitigation methodologies and careful human moderation, can help to remove biases and prevent safety issues during deployment.

\section*{Acknowledgements} We thank Yejin Choi for her helpful feedback. JH partially completed this work while at AI2.

%% file: a2_theory.tex
\subsection{Comparison of Implicit Forecasting Approaches in Inference and Uncertainty Tuning}
\label{sec:if_comp}
When doing implicit forecasting with language models, we find two main approaches in the existing literature (i.e., on calibration to correctness). As in the main text, one can retain the whole token distribution during inference. Recall, this is written:
\begin{equation}
\label{eqn:if_full_token}
    \hat{P}_\texttt{IF} = \frac{e^{Z_\texttt{yes} / \tau}}{\sum_{t \in \mathcal{T}} e^{Z_t / \tau}}.
\end{equation}
On the other hand, one can consider a \textit{normalized} probability, restricted to a set of candidate answers. For example, \citet{jiang2021can} suggest this approach for smaller models, lacking appropriate instruction following capability. In our context, the approach is described:
\begin{equation}
\label{eqn:norm_if}
    \hat{P}_\texttt{IFN} = \frac{e^{Z_\texttt{yes} / \tau}}{e^{Z_\texttt{yes} / \tau} + e^{Z_\texttt{no} / \tau}}.
\end{equation}
Primarily, \citet{jiang2021can} argue for this strategy to cope with the spreading of probability across the many different ways to indicate ``yes'' or ``no'', dependent on the question. In modern instruction-following models tuned with human feedback, \citet{kadavath2022language} consider both approaches. In their notation, ``P(True)'' takes the former approach (using the whole distribution), while ``P(IK)'' tunes a classification head on top of the model's internal feature representation, making ``P(IK)'' equivalent to tuning in the fashion of \citet{jiang2021can}. As we are aware, there has not been much exploration of tuning \textit{and} inference in the fashion of Eq.~\eqref{eqn:if_full_token}. We provide a first, formal comparison of these approaches next.
\paragraph{Qualitative Pros \& Cons} We view the approach of Eq.~\eqref{eqn:if_full_token} to be preferable for a few reasons.
\begin{enumerate}
    \item \textbf{Compatibility with Other Tasks:} Most other language modeling tasks require retention of the full token distribution. While we do not experiment with this in our paper, uncertainty tuning of implicit forecasts, as in Eq.~\eqref{eqn:if_full_token} and \S~\ref{sec:sft}, can theoretically be coupled with other tasks in a more broadly scoped fine-tuning pipeline. In contrast, use of normalized probabilities during uncertainty tuning, as in Eq.\eqref{eqn:norm_if}, would require distinct loss functions and inference protocols to be coupled with other tasks (due to differences in score normalization).
    \item \textbf{Generalized Extension:} Extension beyond yes/no answers (and dialogue forecasting) is far easier when using Eq.~\eqref{eqn:if_full_token}. Indeed, we simply change the token in the numerator and add more data instances during training. In the alternative Eq.~\eqref{eqn:norm_if}, we may require additional algorithms/compute to select candidate sets -- as is done by \citet{jiang2021can}. For one, these added protocols can inhibit inference due to compounding errors, and related to our first point, these added protocols also make uncertainty tuning on many different types of tasks and data prohibitive. 
\end{enumerate}
\paragraph{Theoretical and Empirical Characterization of Differences}
We can also consider the difference between these approaches more precisely. We begin with a theoretical result that is both conceptually informative and useful for empirical study. Primarily, we bound the absolute difference between the two types of implicit forecasts. Define $\mathcal{X} = \mathcal{T} - \{\texttt{yes}, \texttt{no}\}$, then:
\begin{equation}
\begin{split}
& \lvert \hat{P}_\texttt{IF} - \hat{P}_\texttt{IFN} \rvert 
= \hat{P}_\texttt{IFN} \cdot \lvert 1 - \hat{P}_\texttt{IF} / \hat{P}_\texttt{IFN} \rvert 
= \hat{P}_\texttt{IFN} \cdot \Bigg \lvert 1 - \frac{1}{\hat{P}_\texttt{IFN} + (1 - \hat{P}_\texttt{IFN}) + \frac{\sum_{x \in \mathcal{X}} e^{Z_x / \tau}}{e^{Z_\texttt{yes} / \tau} + e^{Z_\texttt{no} / \tau}}}\Bigg \rvert \\
& = \hat{P}_\texttt{IFN} \cdot \Bigg \lvert 1 - \Bigg ( 1 + \frac{\sum_{x \in \mathcal{X}}e^{Z_x / \tau}}{e^{Z_\texttt{yes} / \tau} + e^{Z_\texttt{no} / \tau}} \Bigg )^{-1}\Bigg \rvert \\ 
& = \hat{P}_\texttt{IFN} \cdot \Bigg \lvert \frac{\sum_{x \in \mathcal{X}}e^{Z_x / \tau}}{e^{Z_\texttt{yes} / \tau} + e^{Z_\texttt{no} / \tau}} \cdot \Bigg ( 1 + \frac{\sum_{x \in \mathcal{X}}e^{Z_x / \tau}}{e^{Z_\texttt{yes} / \tau} + e^{Z_\texttt{no} / \tau}} \Bigg )^{-1} \Bigg \rvert
\leq \frac{\sum_{x \in \mathcal{X}}e^{Z_x / \tau}}{e^{Z_\texttt{yes} / \tau} + e^{Z_\texttt{no} / \tau}} \\
& \leq \frac{\sum_{x \in \mathcal{X}}e^{Z_x / \tau}}{e^{Z_\texttt{no} / \tau}}
= \sum_x \Bigg ( \frac{e^{Z_x}}{e^{Z_\texttt{no}}} \Bigg )^{1 / \tau} = \sum_x \varepsilon_x^{1 / \tau} 
\end{split}
\end{equation}
where we define $\varepsilon_x = e^{Z_x} / e^{Z_\texttt{no}}$ as the \textit{excess} score ratio for the token $x$. We estimate this value empirically for temperature $\tau = 1$ across all different datasets, splits, and setups for our fine-tuned models, finding a small average of 1.1 (on a 100pt scale). 
\paragraph{Interpretation} So, for $\tau=1$, $\hat{P}_\texttt{IF}$ and $\hat{P}_\texttt{IFN}$ are \textit{practically equivalent} forecasts. The \textit{main difference is the qualitative benefits of $\hat{P}_\texttt{IF}$} we just discussed. Because $x^p$ decreases as a function of $p > 1$ (and $x < 1$), we also know this number will not get larger for smaller $\tau$ -- so, our interpretation for $\tau < 1$ remains the same. On the other hand, our observed difference does grow with $\tau$, which can mar our interpretation if one uses post-hoc correction techniques like $\tau$ scaling. In our experimental study, this is largely unimportant, since our experiments on fine-tuned models actually observe the implicit signal at $\tau = 1$ \textit{before} doing our proposed (simulated) correction technique in \S~\ref{sec:post-process}. 

All this is to say, if one conducts correction as in \S~\ref{sec:post-process}, the tuned forecasting approaches are equivalent with about 1\% difference on average. Consequently, it makes sense to use Eq.~\eqref{eqn:if_full_token}, recouping the potential qualitative benefits discussed earlier. As for use of other correction techniques, we discuss these next.
\subsection{Comparison of Post-Hoc Correction Techniques}
\label{sec:comp_post_hoc}
In this section, we discuss other choices of post-hoc correction, comparing them to our proposal in \S~\ref{sec:post-process}. As we are the first to study language modeling of uncertainty in a conversational forecasting domain, we focus on studies calibrating uncertainty to model correctness. Namely, we consider approaches by
\begin{enumerate}[nolistsep]
    \item \citet{kadavath2022language}, who infer an implicit signal and conduct temperature scaling using the language model's entire predicted token distribution;\footnote{Recall, this is their methodology for ``P(True)''. Their methodology for ``P(IK)'' is most similar to \citet{jiang2021can}.}
    \item \citet{jiang2021can}, who infer implicit signals and scale temperature on only a set of candidate tokens;
    \item and \citet{tian2023just}, who study direct signals of model correctness and briefly suggest their own temperature scaling approach for this their experiments. 
\end{enumerate}
Our approach is mixed. We (a) infer implicit signals using the full token distribution like \citet{kadavath2022language}, and then (b) conduct an ``approximate'' Platt scaling \citep{platt1999probabilistic} on only a set of candidate tokens. Recall, the ``approximation'' (or simulation) in step (b) is what allows the method to be applicable to \textit{both} implicit forecasts and direct forecasts.
% generalizing the proposals of \citet{jiang2019fantastic} and \citet{tian2023just}. 

As a standalone property, this unification is nice. It reduces the computational overhead to study many different methods, since we can use the same inference and post-processing code for all forecasts in our experiments. In addition, we point out our approach is \textbf{computationally efficient}. Indeed, using the approach we propose, correction is done using the probability of the \texttt{yes} token only, so we need only conduct one forward pass and save this single float per instance. In contrast, temperature scaling using the entire token distribution -- for implicit forecasting, as done by \citet{kadavath2022language} -- requires an \textit{increase in forward passes at least proportional to the number of temperatures we try}, or otherwise, about 32000$\times$ more memory to avoid re-computing forward passes by remembering the token scores.

Next, we conduct some theoretical and empirical analyses comparing our correction technique to that of \citet{jiang2021can}, showing it is practically equivalent. Later on, we also compare our post-processing with \citet{kadavath2022language} on implicit forecasts and \citet{tian2023just} on direct forecasts, using an empirical study. Again, we find ours to be practically equivalent (or better on average).
% \paragraph{Avoidance of Hiding Model Errors} For non fine-tuned models, our argument is more conceptual. In particular, much of the motivation behind using candidate sets to estimate implicit signals \citep{jiang2021can} is that they can help to cope with the ``spreading'' of probability across the many ways to indicate an answer. In our context, for pre-trained models, we use system prompts to directly request a specific output format, expecting that large modern models will follow our instructions as designed \citep{ouyang2022training}. Thus, ``spreading'' probability is actually a model failure we would like to capture. Otherwise, we would require more algorithms/compute to select the best candidate sets -- as is done by \citet{jiang2021can} -- which returns us to our previous argument on the computational fairness of the full forecasting algorithm. Note that, for ``good'' pre-trained models -- that heed the system prompt and don't spread probability --, all the theoretical analyses we use to compare implicit signals and post-processing methods will still apply.
% \subsubsection{Formal Comparison of Post-Processing Approaches}
\paragraph{Comparison of Proposed Correction to that of \citet{jiang2021can}}
Recall our ``estimated logit'' post-processing technique described in Eq.~\eqref{eqn:post-process}. We have processed score:
\begin{equation}
    \hat{Z}_\texttt{yes} = \ln \big ( \hat{P} / (1 - \hat{P}) \big ) / \tau - \beta
\end{equation}
and the post-processed forecast $\hat{P}_\texttt{PP} = \hat{P}_\texttt{new}$, expanded as below:
\begin{equation}
\begin{split}
& \hat{P}_\texttt{PP} = (1 + \exp(-\hat{Z}_\texttt{yes}))^{-1} 
= \Bigg ( 1 + e^{\beta}\Big( \frac{1-\hat{P}}{\hat{P}}\Big)^{1/\tau}\Bigg )^{-1}
= \frac{\hat{P}^{1 / \tau}}{\hat{P}^{1 / \tau} + e^{\beta} (1 - \hat{P})^{1 / \tau}} \\
& = \frac{e^{Z_\texttt{yes} / \tau}}{e^{Z_\texttt{yes} / \tau} + e^{\beta} (e^{Z_\texttt{no}} + \sum_x e^{Z_\texttt{x}})^{1 / \tau}}
\end{split}
\end{equation}
Defining $\varepsilon = \sum_x \varepsilon_x$ we have
\begin{equation}
\label{eqn:bound}
\begin{split}
&\lvert \hat{P}_\texttt{PP} - \hat{P}_\texttt{IFN} \rvert
= \Big \lvert \frac{e^{Z_\texttt{yes} / \tau}}{e^{Z_\texttt{yes} / \tau} + e^{\beta}(1 + \varepsilon)^{1 / \tau} e^{Z_\texttt{no} / \tau}} - \frac{e^{Z_\texttt{yes} / \tau}}{e^{Z_\texttt{yes} / \tau} + e^{Z_\texttt{no} / \tau}}\Big \rvert \\
& = e^{Z_\texttt{yes} / \tau} \cdot \Big \lvert \frac{e^{Z_\texttt{no} / \tau} - e^{\beta}(1 +\varepsilon)^{1 / \tau}e^{Z_\texttt{no} / \tau}}{e^{2Z_\texttt{yes} / \tau} + e^{Z_\texttt{yes} / T + Z_\texttt{no} / \tau} + e^{\beta}(1 +\varepsilon)^{1 / \tau}e^{Z_\texttt{yes} / T + Z_\texttt{no} / \tau} + e^{\beta}(1 +\varepsilon)^{1 / \tau}e^{2Z_\texttt{no} / \tau}} \Big \rvert \\
& = \Big \lvert \frac{1 - e^{\beta}(1 +\varepsilon)^{1 / \tau}}{e^{Z_\texttt{yes} / T - Z_\texttt{no} / \tau} + 1 + e^{\beta}(1 +\varepsilon)^{1 / \tau} + e^{\beta}(1 +\varepsilon)^{1 / \tau}e^{Z_\texttt{no} / T - Z_\texttt{yes} / \tau}} \Big \rvert \leq \lvert 1 - e^{\beta}(1 + \varepsilon)^{1 / \tau} \rvert
\end{split}
\end{equation}
% Using a Taylor expansion for the last term around $0$, we have for some $0 \leq \xi \epsilon$
% \begin{equation}
%    \lvert \hat{P}_\texttt{PP} - \hat{P}_\texttt{IFN} \rvert \leq 1 - (1 + \varepsilon)^{1 / T} = \frac{\varepsilon^2(T-1)}{2T^2} - \frac{\varepsilon}{T} - \frac{(T-1)(2T-1)\xi^3}{6T^3} \leq \frac{\varepsilon^2(T-1)}{2T^2} - \frac{\varepsilon}{T}.
% \end{equation}
Since $\beta$ is a free parameter, it can always be chosen to be 0 during post-processing (e.g., if we determine it is not helpful based on validation data). Thus, since any deviation due to $\beta$ would be by choice to improve our forecasts, we consider estimation of this bound when $\beta = 0$. Indeed, we can easily collect statistics on $(1 + \varepsilon)$ during the forward passes of our experiments, and do so, estimating this bound value with different selections of $\tau$ in Table~\ref{tab:bound}. We find all differences to be practically negligible.
\begin{table}[]
    \centering
    \begin{tabular}{c|ccccccc}
         $\tau$      & 0.25 & 0.5 & 1 & 1.5 & 1.75 & 2 & 2.5 \\\hline
         Prob. Difference Bound  & 4.3 & 2.1 & 1.1 & 0.7 & 0.6 & 0.5 & 0.4
    \end{tabular}
    \caption{Estimation of bound in Eq.~\eqref{eqn:bound} from data with $\beta = 0$. Note, we report the bound on a 100pt scale.}
    \label{tab:bound}
\end{table}
\paragraph{Comparison of Proposed Correction to that of \citet{kadavath2022language} and \citet{tian2023just}}
Lacking theoretical analysis, we compare our unified post-processing approach to the discussed techniques of \citet{kadavath2022language} and \citet{tian2023just}, empirically. To keep all methods on equal footing, we use our fine-tuned Llama-2 7b model for implicit forecasts; i.e., since the method of \citet{kadavath2022language} only operates on implicit forecasts. We also limit study to one prompt setup (the data inferred prior) because the method of \citet{kadavath2022language} is more computationally expensive. To evaluate, in Table~\ref{tab:comps}, we look at performance at minimizing Brier score on the validation set, reporting the percent of times we would have preferred the method of \citet{tian2023just} or \citet{kadavath2022language} due to a lower Brier score. We also report the average magnitude of preference; i.e., how much smaller the Brier score is. The logic here is to show the utility of each post-processing method in a \textit{practical setting}, since we would never actually select a method that does worse on validation data in practice.
%\footnote{This evaluation ignores possible over-fitting, but most methods select only 1 or 2 parameters, so we consider this risk a minimal one.} 
Indeed, the results show that if we \textit{did} use these other proposals, in conjunction with our own, and picked the best technique for each instance based on validation data, we would have still have used our own proposal most of the time.\footnote{We used only our approach for experiments to avoid extra computational overhead, and because it was best overall.} Moreover, the actual magnitude of preference for other methods is very small, so even when another method is preferred, we would ultimately expect similar performance when transferring to a test set.
% Importantly, it also keeps us from peeking at the test set before making our post-processing decisions. In any case, assuming enough validation data, good validation performance will translate to the test set, especially since most methods are only select a few parameters.
\begin{table}[]
    \centering
    \begin{tabular}{c|cc}
        
        Post-Processing Method & \% Preferred (over ours) & Magnitude (Brier score gain; 100pt)\\\hline
        \citet{kadavath2022language} & 4.2 & 0.03\\
         \citet{tian2023just} & 16.7 & 0.03 \\
         \citet{tian2023just} + Bias Correction & 20.1 & 0.03 \\ \hline
    \end{tabular}
    \caption{Comparison of post-processing methods to our approach. For the method of \citet{tian2023just}, we also consider adding a (new) second parameter -- a bias correction term similar to $\beta$ -- to make it more competitive. We report percent of times each method is preferred (compared to our method) based on validation data, as well as average magnitude of preference. In practice, when validation data is used to select among methods, our technique is generally preferred over existing techniques. Even when others are preferred, the degree of preference is small. Meanwhile, when preferred, our post-processing method has preference magnitude 6$\times$ larger on average (0.19).}
    \label{tab:comps}
\end{table}
\paragraph{Computational Details} To select hyper-parameter $\tau$ and $\beta$, we consider two cases: $\beta \neq 0$ and $\beta = 0$. For the first  ($\beta \neq 0$), we fit a 2 parameter logistic model of the outcome variable (i.e., this optimizes a proper scoring function -- the log score). For the latter case ($\beta = 0$), we use either traditional temperature scaling (optimizing Brier score) or a newer scaling approach called \textit{Expectation Consistency}, as described by \citet{clarte2023expectation}. To decide which of these cases (and subsequent optimization procedures) to use, we compare Brier scores, picking the method that yields the overall lowest on validation data. To re-implement the approach of \citet{tian2023just}, we optimize the functional form $\exp(\log(P) / \tau)$, or more generally $\exp(\log(P) / \tau) / \beta$ when applying bias correction, picking parameters to minimize Brier score; this preserves the authors' primary suggestion that scaling be done so the result is proportional to $p^\alpha$ for some $\alpha$.\footnote{We inferred no other restrictions or implementation details from their description.} In general, we use the \texttt{scipy} optimization package or \texttt{scikit-learn} to implement the aforementioned parameter selection. When re-implementing the correction approach of \citet{kadavath2022language}, it is too computationally costly to use the \texttt{scipy} optimization package, so we conduct a simple linear search for $\tau \in \{0.1, 0.25, 0.5, 0.75, 1, 1.25, 1.5, 1.75, 2\}$.
\clearpage
\subsection{An Off-Policy Policy Gradient Theorem}
\label{sec:pgt_pf}
In this section, we show the claimed result from the main text:
\begin{equation}
\begin{split}
    \nabla_\theta \mathbf{E} [R] = \mathbf{E} \Big [ & \mathsf{s}_{\tilde{T}} \cdot \tfrac{\mu_T(\tilde{T})}{\mu_{\tilde{T}}(\tilde{T})}\cdot \nabla_\theta \log \mu_T(\tilde{T}) \Big ] \\
    \text{where} \quad & T \sim \mathtt{LM}^*_\theta \circ \Phi \circ D, \\
    & \tilde{T} \sim \mathtt{Ref}^* \circ \Phi \circ D, \\
    & \text{and} \ \mathsf{s}_{\tilde{T}} = -\mathsf{s}(\mathsf{p} \circ \tilde{T}, O).
\end{split}
\end{equation}
Let all random variables be as above and fix the mass functions $\mu_T$ and $\mu_{\tilde{T}}$. Then, we have
\begin{equation}
\begin{split}
& -\nabla_\theta \mathbf{E} [R] = \nabla_\theta \mathbf{E} \Bigg  [ \ \sum_{t \in \mathcal{T}^*}\mathsf{s}(\mathsf{p} \circ t, O) \cdot \mu_T(t)\Bigg ] = \mathbf{E} \Bigg  [ \ \sum_{t \in \mathcal{T}^*} \mathsf{s}(\mathsf{p} \circ t, O) \cdot \nabla_\theta \ \mu_T(t)\Bigg ] \\
& = \mathbf{E} \Bigg  [ \ \sum_{t \in \mathcal{T}^*} \mathsf{s}(\mathsf{p} \circ t, O) \cdot \mu_T(t) \cdot \nabla_\theta \ln \mu_T(t)\Bigg ] \\
& = \mathbf{E} \Bigg  [ \ \sum_{t \in \mathcal{T}^*} \mathsf{s}(\mathsf{p} \circ t, O) \cdot \mu_T(t) \cdot \tfrac{\mu_{\tilde{T}}(t)}{\mu_{\tilde{T}}(t)} \cdot \nabla_\theta \ln \mu_T(t)\Bigg ] \\
& = \mathbf{E} \Bigg  [ \ \sum_{t \in \mathcal{T}^*} \mu_{\tilde{T}}(t)\Big ( \mathsf{s}(\mathsf{p} \circ t, O) \cdot \tfrac{\mu_T(t)}{\mu_{\tilde{T}}(t)} \cdot \nabla_\theta \ln \mu_T(t)\Big ) \Bigg ] \\
& = \mathbf{E} \Bigg  [ \mathsf{s}(\mathsf{p} \circ \tilde{T}, O) \cdot \tfrac{\mu_T(\tilde{T})}{\mu_{\tilde{T}}(\tilde{T})} \cdot \nabla_\theta \ln \mu_T(\tilde{T}) \Bigg ].
\end{split}
\end{equation}
So, we have our desired result.
\subsection{Off-Policy Implementation Details}
\label{sec:pgt_detail}
Next, we'll discuss some choices for the reference policy $\mathtt{Ref}$. The formal framework we've provided actually allows us to recover variants of some existing techniques for getting language models to output forecasts in token space.
\paragraph{The Quantizer} \citet{lin2022teaching} fine-tune an LM to forecast the correctness of its answers in token space for a factual question-answering task. Primarily, they use the accuracy of different question types to assign confidence levels that the LM should predict for each type. We propose to extend this idea to more general settings via clustering. Instead of assuming pre-assigned partitions, we infer the partitions by clustering the data. The average outcome of a cluster is computed and assigned to each datum in the cluster, which defines a deterministic reference policy $\mathtt{Ref}^*$ to be used in Eq.~\eqref{eqn:policy-opt}:
\begin{equation}
    \mathtt{Ref}^*_\mathtt{C}(X) = \mathsf{p}^\dagger \Big [ \lvert C(X)\rvert^{-1} \sum_{N \in C(X)} O_N \Big ]
\end{equation}
where $C(X)$ is the neighborhood of $X$, $O_N$ is the outcome of neighbor $N$, and $\mathsf{p}^\dagger : [0,1] \to \mathcal{T}^*$ is an inverse for $\mathsf{p}$,\footnote{$\mathsf{p}$ is not bijective in general, but we can consider a subset of $\mathcal{T}^*$ -- e.g., strings like ``72\%'' -- for which $\mathsf{p}^\dagger$ does exist.} mapping probabilities to tokens. In experiments, $C$ is defined by $k$-means clustering over the internal feature representations of $\mathtt{LM}_\theta$. These representations are the average (over time) of the last hidden layer of the model, ignoring masked inputs, and they are updated each epoch when clusters are re-assigned. In practice, we pick $k$ by hyper-parameter tuning and run $k$-means \textit{individually} for each dataset, re-aggregating the cluster assignments afterwards; our motivation for this is to prevent uninformative, imbalanced cluster assignments, which may occur if clusters correlate with dataset labels.
\paragraph{The Exploiter} Given any pre-trained, fixed implicit signal forecaster, we can use it to train a direct forecaster via Eq.~\eqref{eqn:policy-opt}. For example, assuming we have trained a LM via supervised fine-tuning (\S~\ref{sec:sft}) and fixed its forecasting function $\hat{P}_\texttt{IF}$, we define the deterministic reference policy:
\begin{equation}
    \mathtt{Ref}^*_\texttt{S}(X) = \mathsf{p}^\dagger \big [\hat{P}_\texttt{IF} \big ].
\end{equation}
This provides a nice controlled view for the differences between implicit and direct forecasting, since the direct policy is actually learning from the implicit policy. Properties of the implicit policy that do not transfer to the direct policy will be of interest. As noted in the main text, this also represents a focus on exploitation, since the implicit forecasts $\hat{P}_\texttt{IF}$ were designed to maximize the log-likelihood on the training data. Maximizing the log-likelihood is equivalent to minimizing the negative log-likelihood (i.e., the log score) and it is known that the log score is a proper scoring function.
\paragraph{The Explorer} Finally, it is interesting to consider that Eq.~\eqref{eqn:policy-opt} indicates the language model $\mathtt{LM}_\theta^*$ can learn from any policy, even a bad one. To explore this, we suggest a context-less binomial reference policy which simply assigns random probability estimates to the dialogues. Presumably, by Eq.~\eqref{eqn:policy-opt}, $\mathtt{LM}_\theta^*$ can observe the rewards from these estimates and begin to make ``sense'' of them. For binomial parameters $n$ (number of trials) and $\pi$ (success ratio) we define the reference policy:
\begin{equation}
    \mathtt{Ref}^*_\texttt{B}(X) = \mathsf{p}^\dagger \big [B \big ]; \quad B \sim \mathrm{Bin}(n, p).
\end{equation}
In experiments, $n = 20$ and $p$ is the average outcome in the training data (one for each dataset).
\paragraph{On Policy} Alternatively, we can actually use $\mathtt{LM}_\theta^*$ as its own reference: $\mathtt{Ref}^*_\texttt{PPO} = \mathtt{LM}_\theta^*$. As alluded by our notation, this makes Eq.~\eqref{eqn:policy-opt} equivalent to on-policy policy gradient techniques, like Proximal Policy Optimization \citep{schulman2017proximal}. We leave investigation of on policy learning to future work.

%% file: a1_results.tex
\begin{table*}[]
\centering
\begin{tabular}{l|lll|llll}
& \multicolumn{3}{c}{\# Train/Test Matches} & \multicolumn{4}{c}{\# Test Sets Matching Majority of Train}\\
\textbf{Split} & \textbf{Topic} & \textbf{Topic+L} & \textbf{Outcome} & \textbf{Length} & \textbf{Affective} & \textbf{Non-Affective} & \textbf{Multi-party} \\ \hline
\textit{easy}        & all         & all & 1/2                       & all                & all      & all       & all                  \\
\textit{medium}      & 2/3        & none & none                     & all                 & 2/3 & none       & all                  \\
\textit{hard}        & none        & none & none                     & none                  & none & 2/3          & 1/3                 
\end{tabular}
\caption{Table cells show count of test sets that \textit{share} properties with the train set. \textit{easy} has the most shared properties (least imbalance), while \textit{hard} has the least shared properties (most imbalance). We operate under the assumption that greater degrees of imbalance correlate with greater degrees of difficulty in generalization, which is consistent with the domain generalization literature \citep{gulrajani2020search}. Generally, by design, \textit{imbalance of shared properties} (between train and test) \textit{increases} from \textit{easy} to \textit{hard}, creating a sliding scale of difficulty for the generalization problems we study. The heldout datasets for each split are listed in Table~\ref{tab:splits}. The first three columns (\textbf{Topic}, simultaneous \textbf{Topic} + \textbf{L}ength, and \textbf{Outcome}) display how many of the test datasets in the split share the column property with at least one of the train datasets. For example, in \textit{easy} 1 out of 2 test datasets share an \textbf{Outcome} with a train dataset in the same split. The remaining columns (\textbf{Length} and so on) show how many of the test datasets in each split share the column property with \textit{a majority} of the training datasets in the same split. For example, in \textit{hard} 1 out of 3 test datasets has an \textbf{Affective} outcome and the majority of the train datasets have a \textbf{Non-Affective} outcome (causing the ``none'' designation for this column). When comparing length and multi-party similarities, we assume it is easier to generalize from long to short data or multi- to single-party data. So, ``sharing'' means being \textit{at least as long} or having \textit{at least as many parties}.} 
\label{tab:imbalance}
\end{table*}
\begin{table*}[]
\centering
\begin{tabular}{l|lll|lll}
& \multicolumn{3}{c}{Matching Train Set} & \multicolumn{3}{c}{\# Matching Train Sets} \\
\textbf{Split / Heldout Set} & \textbf{Topic}
& \textbf{Topic+L} & \textbf{Outcome} & \textbf{Length}    & \textbf{Affective}    & \textbf{Multi-party} \\ \hline
\textit{easy} /  wiki. (attack)       & deleted        & deleted  & reddit     & 4/6 long     & 3/6 yes     & 3/6 yes            \\
\textit{easy} / item allocation       & camp         & camp   & none     & -             & 3/6 no & -                     \\ \hline
\textit{med} / craigslist   & item alloc. & none   & none       & -             & 2/5 no & -                     \\
\textit{med} / wiki. (deleted)     & none           & none   & none      & 4/5 long     & 2/5 no & 3/5 yes              \\
\textit{med} / camp provisions      & item alloc.          & none   & none      & -             & 3/5 yes     & -                      \\ \hline
\textit{hard} / courtroom     & none           & none   & none      & 2/5 long & 3/5 no & 2/5 yes              \\
\textit{hard} / charity   & none           & none    & none     & 2/5 long     & 2/5 yes     & -                      \\
\textit{hard} / reddit         & none           & none  & none       & 2/5 long     & 2/5 yes     & 2/5 yes             
\end{tabular}
\caption{This table ``shows our work'' for the assertions of imbalance in Table~\ref{tab:imbalance}. It provides a description of similarities and dissimilarities between train and test sets when each dataset is one of those heldout in the split. For each heldout test dataset, the first three columns show similarity in topic, simultaneous topic + length, noting the \textit{specific} train dataset that shares this property. The next three columns show the \textit{number} of similar datasets among the training data, considering length, usefulness of affective reasoning, and presence of multi-party dialogues. Length ``long'' is categorized by having more than 2K characters on average. Recall, when comparing length and multi-party similarities, we assume it is easier to generalize from long to short data or multi- to single-party. So, we put dashes in for ``short'' or single-party data to note imbalances need not be measured.}
\label{tab:splits}
\end{table*}
\subsection{Additional Dataset Details}
\label{sec:data_details}
When available, we use default train/val/test splits from each dataset's original proposal paper. When not available, we split each datatset according to a 70/15/15\% split. All numbers reported in the main text are computed on the unseen test set for each individual dataset (i.e., even for the \textit{in domain} setting).
\paragraph{Sampling of Training Data} For every epoch of training, we sample 750 dialogues from each of the heldin training datasets of the current training split (i.e., \textit{easy}, \textit{medium}, or \textit{hard}). We pick 750 because this ensures a balanced sample across the training data (all datasets have at least 750 training dialogues). Each dialogue is then randomly truncated to $K \sim \mathcal{U}\{2\ldots L\}$ turns where $L$ is the original dialogue's turn length. As we generally train for 5 epochs, this means the model sees roughly 19-23K partial dialogues, with some dependency in examples across epochs (for the smaller datasets). We also tried using larger, imbalanced data samples for training. In this case, we sample 5K dialogues, or as many as are available. Accounting for the datasets that have less than 5K training examples, we estimate the model sees about 4$\times$ more data overall.
\paragraph{Sampling of Test and Validation Data} For each dataset, we use the same validation and test sets across all experiments. We sample 250 dialogues from the validation split of each dataset, and randomly truncate each (in exactly the same way). We sample 550 dialogues from the test split of each dataset, again, randomly truncating each in the same way for all experiments. Some datasets have less than 550 dialogues total in their test split, in which case we use all of the available test dialogues.
\subsection{Hyper-Parameters and Prompts}
\label{sec:model_details}
\paragraph{Hyper-Parameters and other Training Details} Generally, we train for 5 epochs with a batch size of 12 using AdamW for optimization \citep{DBLP:journals/corr/abs-1711-05101}. We use 4bit QLoRA \citep{dettmers2023qlora} with LoRA rank 32. On 4 NVIDIA RTX A6000 GPUs, single model training is an overnight process, so we only conduct full hyper-parameter selection (linear search) on the \textit{medium} split using Llama-2-chat 7B to save time. We use the best hyper-parameters for Llama-2 7B on \textit{medium} for all other train/test splits and models. For implicit forecast tuning, we pick the learning rate from the range \{1e-4, 2e-5, 1e-5\}. For direct forecast tuning, we pick the clipping constant $\epsilon$ from the range \{0.2, 0.5, 0.8\} and the learning rate from the smaller range \{1e-4, 1e-5\} to save time. Clustering for the Quantizer off-policy is also selected from the range \{10, 20\}. Log score on the in-domain validation data is used to pick the best parameters.\footnote{In domain generalization, its important to avoid picking parameters using the held out domains, since this can bias results \citep{gulrajani2020search}.} Parameter selection is fairly consistent overall, with most tuning setups preferring the highest learning rate. $\epsilon$ was always 0.5 and the number clusters for the Quantizer off-policy was 10. 
\paragraph{Inference Parameters} As noted in our discussion, implicit forecasting uses $\tau = 1$ in Eq.~\eqref{eqn:if_full_token} conducting post-hoc correction using our ``estimated logit'' procedure, as in Eq.~\eqref{eqn:post-process}, after the fact. For sampling, to conduct direct forecasting, we typically use the default hyper-parameters indicated by the model parameters (e.g., in the API, Huggingface generation configuration, or Github repository). For GPT-4 this means temperature and top p are both set to 1. For Llama-2 models, this means temperature and top p are set 0.6 and 0.9, respectively. The only exceptions are the pre-trained Llama-2 7b (no fine-tuning) and the Llama-2 7b with direct forecasts tuned on 4$\times$ more data, which both use greedy sampling due to clear (unsupervised) problems with the forecasts; e.g., output scores that prohibited usual sampling due to NaN values or extreme variance in token samples (200x more than greedy). For models tuned in the Zephyr style, temperature is 0.7 and top p is 1. For direct forecasting with pre-trained models, we sample up to 256 tokens to allow room for an explanation (frequently provided by some models). Tuned models generally followed the training format, so we only needed to sample up to 8 tokens in these cases. Upon sampling, our parser extracted a probability forecast by searching for percentages, ratios, and floats in [0,1] (in that order) using regex matching. Before parsing, some logic is also used to drop any non-answer explanations the model provides, i.e., this is based on keywords specified in the system prompts.
\paragraph{Prompts} As noted in the main text, we use a standard formula for prompting all models. We provide an example of the of prompt format used for tuning Llama-2 7B below (with model tags):
\begin{displayquote}
\texttt{<s>[INST] [Segment Start]}

\texttt{<partial dialogue inserted>}

\texttt{[Segment End]}

\texttt{In the preceding conversation segment, a group of Wikipedia contributors are deciding whether an article should be deleted. On average, this type of conversation ends with the article being deleted about 56\% of the time. What is the percent probability that this specific conversation will end with the article being deleted? [/INST] </s>}
\end{displayquote}
and another example: 
\begin{displayquote}
\texttt{<s>[INST] [Segment Start]}

\texttt{<partial dialogue inserted>}

\texttt{[Segment End]}

\texttt{In the preceding conversation segment, the speakers are defending their opinions on an issue. On average, this type of conversation ends with a personal attack about 50\% of the time. What is the percent probability that this specific conversation will end with a personal attack? [/INST] </s>}.
\end{displayquote}
For pre-trained models, we also use a consistent system prompt to constrain output format and provide more task details. Below is the prompt used for direct forecasts:
\begin{displayquote}
\texttt{You are NegotiationPredictionGPT, an expert language model at predicting the likelihood of outcomes in human language negotiations. You will be given the first part of a conversation between several different speakers with potentially different goals. Use the first part of the conversation to put yourself in the mindset of the speakers and estimate the likelihood of the requested conversation outcome for these particular speakers. Use the keyword "OUTCOME" to report your predicted probability for the outcome of interest, requested by the user. Report the probability as a percentage using the format "OUTCOME = percent". For example, "OUTCOME = 70\%}
\end{displayquote}
We focus on prompts for direct forecasts in these examples, but prompts for implicit forecasts are similar, changing only the main question asked (to evoke a yes/no response).
\section{Examples}
Here, we provide some examples of negotiations the model would see during training and testing.

\noindent A \textbf{wiki. editing} example where the outcome of interest is the occurrence of a personal attack:
\begin{displayquote}
\textit{Speaker 3: Material moved from anon edit for discussion.  I vaguely remember this or a similar incident from a TV news program.  But I thought the kids were older.  Notable?  Verifiable?}

\textit{''In 2005, approximately twenty sixth grade students at Reading Fleming Middle School (now Reading Fleming Intermediate School) in Flemington, New Jersey contracted syphilis after attendeding a "rainbow party".''}

\textit{Speaker 3: Rainbow party}

\textit{Speaker 2: I don't really see much point in reporting every case of syphilis ever reported.... - }

\textit{Speaker 1: Indeed. There is no source, it sounds rather urban-legendlike, and a rainbow party is a sure ingredient for those kind of tales. Sure enough, Googling for the string ''Flemington "rainbow party" syphilis'' gives 0 hits.}

\textit{Speaker 0: Source from Hunterdon Central Regional High School in Flemington, New Jersy
http://central.hcrhs.k12.nj.us/bezsylko/discuss/msgReader\$281?mode=day}

\textit{I don't think a teacher would assign that if it wasn't true and, trust me, it is. One of my friend's sister's was one of the girls who contracted it. So, I'd appreciate it if you didn't accuse it of being an "urban legend".}

\textit{Speaker 1: This is ''absolutely'' not a reliable source, apart from the fact that this received NO media coverage. Please stop reinserting this. When MMWR reports this, we can talk again.}
\end{displayquote}
\newpage
\noindent An example from \textbf{craigslist} where the outcome of interest is whether the buyer will get the best deal:
\begin{displayquote}
\textit{Speaker 0: I am interested in this apartment! Can you tell me more about it?}

\textit{Speaker 1: This apartment is located in San Pablo and close to everything! You will have a short commute to the office, the hottest stores, and the newest restaurants! The apartment has lots of closet space, two bedrooms, large windows that really brighten up the space, and an enclosed patio on the back.}

\textit{Speaker 0: Great. I'm looking for a place in that area. Is a security deposit required?}

\textit{Speaker 1: Right now we have a special . . . \$99 security deposit! But you have to take advantage of the offer today!}

\textit{Speaker 0: Would you be willing to go down to \$800 for the first month's rent?}

\textit{Speaker 1: I am sorry, but the rent is \$1725 . . . \$800 is much too low.}

\textit{Speaker 0: What about \$1,200?}
\end{displayquote}

\noindent An example from the \textbf{charity} discussions where the outcome of interest is a occurence of a donation:
\begin{displayquote}
\textit{Speaker 1: Hi!  Have you heard of an organization called Save the Children?}

\textit{Speaker 0: I think I have once before, in a grocery store I believe}

\textit{Speaker 1: Do you mind if i give you a little information about them?}

\textit{Speaker 0: Sure, go ahead}

\textit{Speaker 1: Just some ver basic info, Save the Children is an international non-governmental organization that promotes children's rights, provides relief and helps support children in developing countries.}

\textit{Speaker 0: Are they a non profit organization?}

\textit{Speaker 1: Yes they are! They work 100\% on donated funds. There is  a lack of support for children in developing countries, especially in war zones. For instance, millions of Syrian children have grown up facing the daily threat of violence. In the first two months of 2018 alone, 1,000 children were reportedly killed or injured in intensifying violence. Your donation can address such problems.}

\textit{Speaker 0: Oh wow, shocking news. Do you know how many children have been helped due to this organization?}

\textit{Speaker 1: According t0 their yearend report they were able to reach 155 million children.  Over 200k of those kids were in the US.}

\textit{Speaker 0: Thats awesome! Are you apart of this organization or just support them?}

\textit{Speaker 1: I am just a supporter but I would like to ask how much do you like to donate to the charity? Your donation will be directly deducted from your task payment. You can choose any amount from \$0 to all your payment}
\end{displayquote}
\newpage
\noindent An example from the \textbf{courtroom} discussions where the outcome of interest is whether the petitioner will have a favorable decision:
\begin{displayquote}
\textit{Speaker 2: I was reading, Your Honor, from the only place that I know of that the findings of fact of the district judge are reprinted.
They're in the petition -- the appendix -- no, Your Honor, that's their brief.
The petition for certiorari, page 45(a) --}

\textit{Speaker 6: 45 (a).}

\textit{Speaker 2: Which includes the district judge's opinion and findings of fact and conclusions on this remand hearing.
Thank you.}

\textit{Speaker 5: Mr. Glasser.}

\textit{Speaker 1: Thank you, Your Honor.
The Kolod-Alderisio problem in our case would exist only in relation to the Florida bugging.
We agree with the government.
There's no real issue on Florida, but there is a very severe issue, we say, in connection with an allegedly abortive additional bugging in Georgia.
I haven't spoken of that today.
We've briefed it pretty completely, and I would ask the Court to watch for that item since there was some animation here at the end about the Kolod problem which I think it is currently before the Court.
Now --}

\textit{Speaker 3: Well, they didn't get -- they didn't make any tapes at all or get any recordings, did they, in that second incident to which you refer?} 

\textit{Speaker 1: They -- the agent who ran it said he didn't get the tape, and I think one other agent who was in the car with him said they didn't get any effective audible results.
But, again, we had a very hard pushing hearing in which I, for one, can wait, feeling that I was entitled to make a strong appellate point against the credibility of those agents on that issue too.
And, indeed, on that issue above all, they were crawling all over that part of Georgia.
They were there about to score, and they were not hesitating to bug.
They were bugging all over the country.
We think we can't prove that they were bugging in Europe.
These fellows lived with bugs.
It's incredible to me that they didn't have more than that one abortive car bug in Georgia.
They must have bugged Desist's room.
I'm speaking of perhaps -- well, all right, I'll drop that point for now because it's been thoroughly briefed.
Our whole submission is sufficiently stated in the briefs.
Now, on Fuller, again, may I say something that is -- have been abrupt.
We think this Court should withdraw its action in Fuller on the ground that certiorari there was improvidently granted and I'd like to say why.
We've covered it thoroughly in our last brief.
Fuller involved a telegram, we all know that, but back of that telegram was a subpoena.
The police in Fuller were not defiant or willful towards existing law.
The police in Fuller went to the Alaska Communications Body, whatever it's called, got voluntary relinquishment of the telegram from that body pursuant to a federal regulation and they also got a subpoena.
Now, the exact details of that whole subpoena picture, I don't know for sure of myself because I haven't seen the Fuller record but I've been guided through it in consultation in clause, consultation with one of the Fuller certiorari counsel.
I have the page numbers.
This is covered in our last brief.
Now, if there was a subpoena in Fuller for that telegram, how can Your Honors reach the question in Fuller of a violation of 605 because the very first sentence of 605 provides for subpoenas nor, at least colorably and subject to a closer scrutiny of the record in Fuller than -- which Your Honors may well wish to do because Fuller is a pretty drastic decision, and to render a drastic decision like Fuller on a record that may not stand up under scholarly criticism one of these days, I would think would be something that the Court that wish to hear about.}

\textit{Speaker 7: What did the Alaska Court held in Fuller -- }

\textit{Speaker 1: The Alaska Court never -- }

\textit{Speaker 7: With respect to the 605 violation?}

\textit{Speaker 1: Never touch this problem that I'm talking about now.
Oh, well, they touched the 605 problem.}

\textit{Speaker 7: What did it involve with respect to the 605 violation?}

\textit{Speaker 1: Yes, they touched the 605, but they didn't touch the problem of subpoena pursuant to 605.}

\textit{Speaker 7: What did the Alaska Court hold with respect to the 605 violation of Fuller?}

\textit{Speaker 1: They held that -- let me think.
Now, wait a minute.}

\textit{Speaker 7: There's a dissent, but the Court held that --}

\textit{Speaker 1: They -- oh, they held that 605 does not apply to states that they adopted the basic Schwartz line.}

\textit{Speaker 7: Yes.}
\end{displayquote}